\documentclass[twoside]{article}
\usepackage[accepted]{aistats2021}
\usepackage[T1]{fontenc}
\usepackage[latin9]{inputenc}
\usepackage{lmodern}
\usepackage{color}
\usepackage{verbatim}
\usepackage{float}
\usepackage{amsmath}
\usepackage{mathtools,array,booktabs}
\usepackage{amsthm}
\usepackage{amssymb}
\usepackage{graphicx}
\usepackage[authoryear]{natbib}
\usepackage{hyperref}
\usepackage[final,tracking=true,kerning=true,spacing=true,factor=1100,stretch=20,shrink=20]{microtype}
\usepackage{algorithm}
\usepackage{algpseudocode}
\usepackage{blindtext}
\makeatletter

\floatstyle{ruled}
\newfloat{algorithm}{tbp}{loa}
\providecommand{\algorithmname}{Algorithm}
\floatname{algorithm}{\protect\algorithmname}

\theoremstyle{plain}
\newtheorem{thm}{\protect\theoremname}

\theoremstyle{plain}
\newtheorem{prop}[thm]{\protect\propositionname}
\theoremstyle{remark}

\theoremstyle{plain}
\newtheorem{cor}[thm]{\protect\corollaryname}
\theoremstyle{plain}
\newtheorem{lem}[thm]{\protect\lemmaname}

\usepackage[table]{xcolor}
\graphicspath{{../figures/}}
\makeatother

\usepackage[english]{babel}
\providecommand{\corollaryname}{Corollary}
\providecommand{\lemmaname}{Lemma}
\providecommand{\propositionname}{Proposition}
\providecommand{\remarkname}{Remark}
\providecommand{\theoremname}{Theorem}

 \belowdisplayskip      \abovedisplayskip

\usepackage[colorinlistoftodos,bordercolor=orange,backgroundcolor=orange!20,linecolor=orange,textsize=scriptsize]{todonotes}

\newcommand{\E}{\mathbb{E}}
\newcommand{\R}{\mathbb{R}}

\newcommand{\cO}{\mathcal{O}}
\newcommand{\SGDM}{\texttt{SGDM} }

\begin{document}

\twocolumn[

\aistatstitle{The Power of Factorial Powers: New Parameter settings for (Stochastic) Optimization}

\aistatsauthor{Aaron Defazio \And Robert M. Gower  }

\aistatsaddress{ Facebook AI Research \\ New York  \And  Facebook AI Research \\New York} ]

\begin{abstract}
The convergence rates for convex and non-convex optimization methods depend on the choice of a host of constants, including step sizes, Lyapunov function constants and momentum constants. In this work we propose the use of factorial powers as a flexible tool for defining constants that appear in convergence proofs. We list a number of remarkable properties that these sequences enjoy, and show how they can be applied to convergence proofs to simplify or improve the convergence rates of the momentum method, accelerated gradient and the stochastic variance reduced method (SVRG).
\end{abstract}

\section{Introduction}
\label{sec:intro}Consider the stochastic optimization problem
\begin{equation}
\label{eq:prob}
x_* \in \arg\min_{x \in C} f(x)=\E_{\xi}\left[f\left(x,\xi\right)\right],
\end{equation}
where each $f\left(x,\xi\right)$ is convex but potentially non-smooth in $x$ and $C \subset \R^d$ is a bounded convex set.
To solve~\eqref{eq:prob} we use an iterative method that at the $k$th iteration samples a stochastic (sub-)gradient $\nabla f(x_k,\xi)$ and uses this
gradient to compute a new, and hopefully improved, $x_{k+1}$ iterate.  The simplest of such methods is \emph{Stochastic Gradient Descent} (SGD) with projection:
\begin{equation}\label{eq:SGD}
x_{k+1} = \Pi_{C} \left ( x_k - \eta_k \nabla f(x_k,\xi) \right),
\end{equation}
where $\Pi_C$ is the projection onto $C$ and $\eta_k$ is a sequence of step-sizes. Both the variance from the sampling procedure, as well as the non-smoothness of $f$ prevent the sequence of $x_k$ iterates from converging. 
The two most commonly used tools to deal with this variance are iterate averaging techniques~\citep{Polyak64} and decreasing step-sizes~\citep{RobbinsMonro:1951}. By carefully choosing a sequence of averaging parameters and decreasing step-sizes we can guarantee
that the variance of SGD will be kept under control and the method
will converge. In this work we focus on an alternative to averaging: momentum.
 Momentum
 can be used as a replacement for averaging for non-smooth problems, both in the stochastic and non-stochastic setting. Projected SGD with momentum can be written as 
\begin{eqnarray}
    m_{k+1} &= &  \beta m_{k} + (1-\beta)\nabla f(x_k, \xi_k)\label{eq:gdmupdate-intro} \nonumber,\\
    x_{k+1} &=& \Pi_{C}\left(x_k -\alpha_k m_{k+1}\right), \label{eq:SGDM}
\end{eqnarray}
where $\alpha_{k}$ and $\beta$ are step-size and momentum parameters respectively. 
Using averaging and momentum to handle variance introduces a new problem: choosing
and tuning the additional sequence of parameters. In this work we introduce the use of factorial powers for
the averaging, momentum, and step-size parameters. As we will show, the use of factorial powers simplifies and strengthens the convergence rate proofs.

\paragraph{Contributions}
\begin{enumerate}
   \item We introduce factorial powers as a tool for providing tighter or more elegant proofs for the convergence rates of methods using averaging, including dual averaging and Nesterov's accelerated gradient method, see row 5 in Table~\ref{tab:results}.
	\item We leverage factorial powers to prove tighter any-time convergence rates for SGD with momentum  in the non-smooth convex and strongly-convex cases, see  rows 1 and 2 in Table~\ref{tab:results}.
	\item We describe a novel SVRG variant with inner-loop factorial power momentum, which improves upon the SVRG++~\citep{SVRG++} method in both the convex and strongly convex case, see rows 3 and 4 in Table~\ref{tab:results}.
	\item We identify and unify a number of existing results in the literature that make use of factorial power averaging, momentum or step-sizes.
\end{enumerate}

\begin{table*}[ht]
\centering \small
\begin{tabular}{|c|c|c|c|c|c|c|} \hline 
Method & Alg \# &  Smooth &  Str. Conv & Polytopic Rate &Std. Rate   & Reference\\ \hline 
\SGDM &  Eq~\eqref{eq:iam-main} & No & No & $(n+2)^{\overline{-1/2}}$ & $(n+1)^{-1/2}$ &   \cite{primal_averaging}\\
\SGDM &  Eq~\eqref{eq:iam-main} & No & Yes & $(n+2)^{\overline{-1}}$ & $(n+1)^{-1}$ &   \cite{primal_averaging}\\
SVRGM&  Alg~\ref{alg:svrg+m} & Yes & No & $1/n$ & $1/n$ &  Allen Zhu \& Yuan [2016]
 \\
SVRGM&  Alg~\ref{alg:svrg+m} & Yes & Yes & $(3/5)^{n/\kappa}$ & $(3/4)^{n/\kappa}$  & Allen Zhu \& Yuan [2016]
 \\
Nesterov & Eq~\eqref{eq:Nesterov}  & Yes & No    & $1/n^{\overline{2}}$ & $1/n^2$ & \cite{Nesterov-convex} \\ \hline
\end{tabular}
\caption{List of convergence results together with previously known results. We say that the function is smooth if~\eqref{eq:smooth} holds with constant $L$, otherwise we assume that the function is $G$--Lipschitz~\eqref{eq:lipschitz}.   Finally  when assuming the function if $\mu$--strongly convex we use $\kappa := L/\mu$. The SVRGM is in fact a new method which is closely related to the SVRG++~\cite{SVRG++} method.}
\label{tab:results}
\end{table*}

\section{Factorial Powers}
\begin{figure*}
\includegraphics[width=0.49\textwidth]{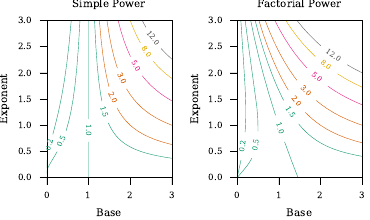}
\includegraphics[width=0.49\textwidth]{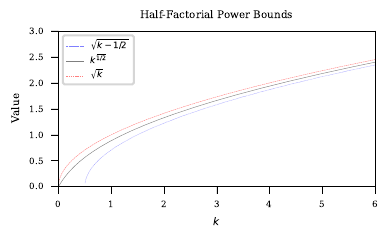}
\caption{(left) Contour plots of the simple powers and the factorial powers. (right) The half-factorial power and associated upper and lower bounds.}
\label{fig:fact}
\end{figure*}
The (rising) factorial powers  \citep{concrete-mathematics} are typically defined using a positive integer $r$ and a non-negative integer $k$ as
\begin{equation}
k^{\overline{r}} \; =\;  k (k+1) \cdots (k+r-1) \; = \; \prod_{i=0}^{r}(k+i-1). \label{eq:poly}
\end{equation}
Their behavior is similar to the simple powers $k^{r}$ as $k^{\overline{r}}=O(k^{r})$, and as we will show, they can typically replace the use of simple powers in proofs. They are closely related to the simplicial polytopic numbers $P_{r}(k)$ such as the triangular numbers $k(k+1)$, and tetrahedral numbers $\frac{1}{6}k(k+1)(k+2)$, by the relation $P_{r}(k)=\frac{1}{r!}$ $k^{\overline{r}}$. See the left of Figure~\ref{fig:fact} for contour plots comparing factorial and simple powers.

The advantage of $k^{\overline{r}}$ over $k^{r}$ is that in many cases that
arise in proofs, additive, rather than multiplicative operations,
 are applied to the constants. As we show in Section~\ref{sec:prop}, summation and difference operations applied to $k^{\overline{r}}$ result in other factorial powers, that is, factorial powers are \emph{closed} under summation and differencing. 
 In contrast, when summing or subtracting simple powers of the form $k^{r}$, the resulting quantities are polynomials rather than
simple powers. It is this closure under summation and differencing that allows us to derive improved convergence rates when choosing step-sizes and momentum parameters based on  factorial powers.

Our theory will use a generalization of the factorial powers to non-integers $r \in \R$
and  integers $k \geq 1$ such that $k+r>0$  using the \emph{Gamma function} $\Gamma(k):=\int_{0}^{\infty}x^{k-1}e^{-x}dx$
so that
\begin{equation}
\boxed{
\;\;k^{\overline{r}}:=\frac{\Gamma(k+r)}{\Gamma(k)}\;\;\label{eq:polygama}
}
\end{equation}
We also use the convention that $0^{\overline{r}}=0$ except for $0^{\overline{0}}=1$. 
This is a proper extension
because, when $k$ is integer we have that $\Gamma(k)=(k-1)!$ and
consequently~(\ref{eq:polygama}) is equal to (\ref{eq:poly}).  This generalized sequence is particularly useful for the values $r=1/2$
and $r=-1/2$, as they may replace the use of $\sqrt{k}$ and $1/\sqrt{k}$
respectively in proofs. 

The factorial powers can be computed efficiently using the log-gamma function to
prevent overflow.
 Using the factorial powers as step sizes or momentum constants adds no computational overhead as they may be computed recursively using simple algebraic operations as we show below.
 The base values for the recursion may be precomputed as constants to avoid the overhead of gamma function evaluations entirely.

\subsection{Notation and Assumptions}
\label{sec:notation} We assume throughout that $f(x,\xi)$ is convex in $x$.  Let $\nabla f(x,\xi_k)$ denote the subgradient of $f(x,\xi_{k})$ given to the optimization algorithm at step $k$.  Let $C \subset \R^d$ be a convex set and let $R>0$ be the radius of the smallest Euclidean-norm ball around the origin that contains  the set $C$. We define the projection onto $C$ as $\Pi_C(x) := \arg\min_{z\in C} \|z-x\| $.
In addition to assuming that $f(x,\xi)$ is convex, we will use one of the following two sets of assumptions depending on the setting.
\vspace{-1em}
\paragraph{Non-smooth functions.}
The function $f(\cdot,\xi)$ is Lipschitz with constant $G>0$
for all $\xi$, that is
\begin{equation}\label{eq:lipschitz}
| f(x,\xi) - f(y,\xi) | \; \leq \; G \; \|x-y\|, \quad \forall x,y \in \R^d.
\end{equation}

\vspace{-1em}
\paragraph{Smooth functions.}
The gradient $\nabla f(\cdot,\xi)$ is Lipschitz with constant $L>0$
for all $\xi$, that is
\begin{equation}\label{eq:smooth}
\| \nabla f(x,\xi) - \nabla f(y,\xi) \| \; \leq \; L \; \|x-y\|, \quad \forall x,y \in \R^d.
\end{equation}
We assume that $\sigma^{2}<\infty$ where
$\sigma^{2}=\E_{\xi}\left\Vert \nabla f\left(x_{*},\xi\right)\right\Vert ^{2}. $

   \vspace{-1em}
\paragraph{Strongly functions.}
We say that $f(x)$ is $\mu$--strongly convex if $f(x)-\frac{\mu}{2}\|x\|^2$ is convex.

We use the shorthand notation $\E_{\xi}\left\Vert \cdot\right\Vert ^{2}=\E_{\xi}\left[\left\Vert \cdot\right\Vert ^{2}\right]$ and
will write $\E$ instead of $\E_{\xi}$ when the conditional context is clear. We defer all proofs to the supplementary material.

\section{Properties of Factorial Powers}
\newcommand{\skipd}{\\[0.2cm]}
\begin{table*}
\centering
\rule{1.0\textwidth}{\heavyrulewidth}
\vspace{-\baselineskip}
\begin{flalign}
&\text{Recursion} & \left(k+1\right)^{\overline{r}} &=\frac{k+r}{k}k^{\overline{r}} &&
\label{eq:recurse_k}\\
& & \left(k+1\right)^{\overline{r}} & =\left(k+r\right)\left(k+1\right)^{\overline{r-1}} &&\label{eq:recurse_r}\skipd
&\text{Summation}
 & \sum_{i=a}^{b}i^{\overline{r}} &=\frac{1}{r+1}b^{\overline{r+1}}-\frac{1}{r+1}\left(a-1\right)^{\overline{r+1}}  &&\label{eq:general_sum} \skipd
&\text{Differences} & \left(k+1\right)^{\overline{r}}-k^{\overline{r}} &=r\left(k+1\right)^{\overline{r-1}} && \label{eq:diffprop}\skipd
&\text{Ratios} & \frac{k^{\overline{r+q}}}{k^{\overline{r}}} &=\left(k+r\right)^{\overline{q}} && \label{eq:ratio}\skipd
&\text{Inversion} &  k^{\overline{-r}}	&= \frac{1}{\left(k-r\right)^{\overline{r}}} && \label{eq:inversion}
\end{flalign}
\vspace{-\baselineskip}
\rule{1.0\textwidth}{\heavyrulewidth}
\caption{Fundamental Properties of the factorial powers. Properties~\eqref{eq:recurse_k},~\eqref{eq:recurse_r},~\eqref{eq:diffprop} and~\eqref{eq:ratio}  hold for $k+r >0$ and $k \geq 1.$  The Summation property~\eqref{eq:general_sum} holds for $a+r >0$ and $a \geq 1.$ The Inversion property~\eqref{eq:inversion} holds for $k > r$ and $k \geq 1.$}\label{tab:prop}
\end{table*}
\label{sec:prop} The factorial powers obey a number of properties, see Table~\ref{tab:prop}. These properties allow for a type of "finite" or "umbral" calculus that uses sums instead of integrals
\citep{concrete-mathematics}. A few of these properties, such as the summation and differencing, are given in Chapter 2.6 for integers values in~\citep{concrete-mathematics}.  We carefully extend these properties to the non-integer setting. All the proofs of these properties can be found in  Section~\ref{sec:properties} in the supplementary material. 

These properties
are key for deriving simple and tight convergence proofs. For instance,  often when using telescoping in a proof of convergence, we often need a \emph{summation} property. For the factorial powers we have the simple formula
~\eqref{eq:general_sum}. This shows that the factorial powers  are \emph{closed} under summation because on both sides of~\eqref{eq:general_sum} we have factorial powers. This formula is a discrete analogue of the definite integral $\int_{a}^{b}x^{r}dx=\frac{1}{r+1}b^{r+1}-\frac{1}{r+1}a^{r+1}$. In contrast, when summing power sequences, we rely on Faulhaber's formula:
\begin{equation}\label{eq:Faulhaber} 
\sum_{i=1}^{k}i^{r}=\frac{k^{r+1}}{r+1}+\frac{1}{2}k^{r}+\sum_{i=2}^{r}\frac{B_{i}}{i!}\frac{r!}{\left(r-k+1\right)!}k^{r-i+1},
\end{equation}
which involves the Bernoulli numbers $B_{j}:=\sum_{i=0}^j\sum_{\nu=0}^i (-1)^v \binom{i}{\nu}\frac{(\nu+1)^j}{i+1}$.  This is certainly not as simple as~\eqref{eq:general_sum}. Furthermore, to extend~\eqref{eq:Faulhaber} to non-integer $r$ complicates matters further~\citep{MCGOWN2007571}. In contrast the summation property
\eqref{eq:general_sum} holds for non-integer values.

Another common property used in telescoping arguments is the \emph{difference} property~\eqref{eq:diffprop}. Once again we have that factorial powers are closed under differencing.
In contrast, the simple powers instead require the use of inequalities such as
\begin{eqnarray*}
rx^{r-1}\leq &\left(x+1\right)^{r}-x^{r} &\leq r\left(x+1\right)^{r-1}
 \\
r\left(x+1\right)^{r-1}\leq & \left(x+1\right)^{r}-x^{r} &\leq rx^{r-1}
\end{eqnarray*}
where the first row of bounds hold for $r <0 $ or $r>1$  and the second row holds for  $r \in (0,1)$.
Using the above bounds adds slack into the convergence proof and ultimately leads to suboptimal convergence rates.

\subsection{Half-Powers}
 The factorial \emph{half}--powers $k^{\overline{1/2}}$ and $k^{\overline{-1/2}}$ are particularly interesting since they can be used to set the learning rate of the momentum method in lieu of the standard $O(1/\sqrt{k})$ learning rate, as we will show  in Theorem~\ref{thm:SGD-nonsmooth-poly}.
The factorial half--powers are similar to the standard half-powers, in that, 
their growth is
 sandwiched by the standard half-powers as illustrated in  Figure~\ref{fig:fact} where we show that
\begin{gather}
\sqrt{(k-1/2)}\leq k^{\overline{1/2}}\leq\sqrt{k}, \\
 \frac{1}{\sqrt{k-1/2}}<k^{\overline{-1/2}}<\frac{1}{\sqrt{k-1}}.
\label{eq:halfbound}
\end{gather}

We also believe this is the first time factorial half--powers have be used in the optimization literature.

\section{From Averaging to Momentum}\vspace{-0.3em}
\label{sec:averaging}\label{sec:momentum}
Here we show that averaging techniques and momentum techniques have a deep connection. We use this connection to motivate the use of factorial power momentum. 
Our starting point for this is SGD with averaging which can be written using the online updating form
\begin{align}
x_{k+1} & =\varPi_{C}\left(x_{k}-\eta_{k}\nabla f(x_{k},\xi_{k})\right),\nonumber \\
\bar{x}_{k+1} & = \left(1-c_{k+1}\right)\bar{x}_{k} + c_{k+1}x_{k+1}.\label{eq:online-average}
\end{align}

At first glance~\eqref{eq:online-average} is unrelated to SGD with momentum~\eqref{eq:SGDM}.  But surprisingly, SGD with momentum can be re-written in the strikingly similar iterate averaging form given by
\begin{align}
z_{k+1} & = \Pi_{C} \left( z_{k}-\eta_{k}\nabla f(x_{k},\xi_{k}) \right) , \nonumber \\
x_{k+1} & =\left(1-c_{k+1}\right)x_{k}+c_{k+1}z_{k+1},
\label{eq:iam-main}
\end{align}
as we prove in the following theorem. This equivalence only holds without the projection operation in Equation~\ref{eq:SGDM}. We are not aware of any analysis of Equation~\ref{eq:SGDM}'s convergence with the projection operation included, and we believe that incorporating projection as we do in Equation~\ref{eq:iam-main} is better given it's much more amendable to analysis.
\begin{thm} \label{theo:equivSGM-IAM}
If $C=\R^d$ then the $x_k$ iterates of~\eqref{eq:SGDM} and~\eqref{eq:iam-main} are the same so long as
$z_0 = x_0$,  $ c_1 \in (0,\,1)$ and
\begin{equation}\label{eq:constsmomtosgdnew}
\eta_{k}=\frac{\alpha_{k}}{c_{k+1}}(1-\beta),\quad c_{k+1}=\beta \frac{\alpha_k}{\alpha_{k-1}} \frac{c_k}{1-c_k}.
\end{equation}
\end{thm}
\begin{proof}
The proof is by induction. 
\paragraph{Base case $k=0$.}
From~\eqref{eq:SGDM} we have that
\begin{align}
x_{1} & = x_0 -\alpha_0 m_{1} \; \overset{\eqref{eq:SGDM}}{=} \;  x_0 - \alpha_0(1-\beta)\nabla f(x_0, \xi_0), \label{eq:mss4o8nsn}
\end{align}
where we used that $m_0 =0.$
Similarly for~\eqref{eq:iam-main} we have that
\begin{eqnarray}
x_{1} & =& \left(1-c_{1}\right)x_{0}+c_{1}z_{1} \nonumber \\
 & = &  \left(1-c_{1}\right)x_{0}+c_{1}(x_0 -\eta_0 \nabla f(x_0, \xi_0)) \nonumber \\
&= & x_0  - c_{1}\eta_0 \nabla f(x_0, \xi_0), \label{eq:sn848pna4}
\end{eqnarray}
where we used that $z_0 = x_0.$ Now~\eqref{eq:sn848pna4} and~\eqref{eq:mss4o8nsn} are equivalent since
$c_1\eta_0 \; \overset{\eqref{eq:constsmomtosgdnew}}{=} \; \alpha_0(1-\beta).$

\paragraph{Induction step.} Suppose that the $x_k$ iterates in~\eqref{eq:online-average} and~\eqref{eq:iam-main}  are equivalent for $k$ and let us consider the $k+1$ step.
Let 
 \begin{equation}\label{eq:zprooftem2p}
  z_{k+1} = x_{k} -\frac{\alpha_k}{c_{k+1}} m_{k+1}.
 \end{equation}
 Consequently
 \begin{eqnarray*}
 z_{k+1} &=& x_{k} -\frac{\alpha_k}{c_{k+1}} m_{k+1} \\
& \overset{\eqref{eq:iam-main}+\eqref{eq:SGDM}}{=}& ( x_{k-1}-\alpha_{k-1} m_{k})  \\ 
& & \quad- \frac{\alpha_k}{c_{k+1}}( \beta m_{k} + (1-\beta)\nabla f(x_{k}, \xi_{k})) \\
&= & x_{k-1}-\left(\alpha_{k-1} +\beta\frac{\alpha_k}{c_{k+1}}\right)m_{k}  \\
 & & \quad - \frac{\alpha_k}{c_{k+1}} (1-\beta)\nabla f(x_{k}, \xi_{k}) \\
&\overset{\eqref{eq:constsmomtosgdnew}}{=} & x_{k-1}-\frac{\alpha_{k-1}}{c_{k}}m_{k} - \eta_k \nabla f(x_{k}, \xi_{k}). \\
&\overset{\eqref{eq:zprooftem2p}}{=} & z_{k}- \eta_k \nabla f(x_{k}, \xi_{k}),
 \end{eqnarray*}
 where in the last but one step we used that  $c_{k+1}~=~\beta \frac{\alpha_k}{\alpha_{k-1}} \frac{c_k}{1-c_k}$ which when re-arranged gives  
 \[ \alpha_{k-1} +\beta\frac{\alpha_k}{c_{k+1}}= \frac{\alpha_{k-1}}{c_{k}}.\]
 Finally
  \begin{eqnarray*}
 x_{k+1} &=& x_k -\alpha_k m_{k+1}\\
 &\overset{\eqref{eq:zprooftem2p}}{=} & x_k  -c_{k+1}(x_{k}-z_{k+1}) \\
& =& (1-c_{k+1}) x_k + c_{k+1} z_{k+1}.
 \end{eqnarray*}
Which concludes the induction step and the proof. \end{proof}

 Due to this equivalence, we refer to~\eqref{eq:iam-main} as the projected \SGDM method.
The $x_k$ update~\eqref{eq:iam-main} is similar to the moving average in~\eqref{eq:online-average}, 
but now the averaging occurs directly on the $x_k$ sequence that the gradient is evaluated on. As we will show, convergence rates of the \SGDM method can be shown for the $x_k$ sequence, with no additional averaging necessary. This method is also known as \emph{primal-averaging}, and under this name it was explored by  \citet{sebbouh2020convergence} in the context of smooth optimization and by \cite{primal_averaging} and \cite{computer-aided2019} without drawing an explicit link to stochastic momentum methods.

Factorial powers play a key role in the choice of the \emph{momentum parameters} $c_{k+1}$, and the resulting convergence rate of~\eqref{eq:online-average}. Standard (equal-weighted) averaging given by
\begin{align}
\bar{x}_k &:= \frac{1}{k+1} \sum_{i=0}^k x_i \quad \mbox{or equivalently} \nonumber \\
 \bar{x}_k & :=\left(1-\frac{1}{k+1} \right)\bar{x}_{k-1} + \frac{1}{k+1} x_k.\label{eq:plain-average}
\end{align}
results in a sequence that ``forgets the past'' at a rate of $1/k$. Indeed, if we choose an arbitrary initial point $x_{0}$ (or at least without any special insight), to converge to the solution we must ``forget'' $x_{0}.$ To forget $x_0$ faster, we can use  a weighted average that puts more weight on recent iterates. We propose the use of the factorial powers to define a family of such weights that allows us to tune how fast we forget the past. In particular, we propose the use of momentum constants as described in the following proposition.
\begin{prop}\label{prop:polydecayisfactorialpower}
Let $x_{k}\in\mathbb{R}^{n}$ for $k=1,\dots$ be a sequence of
iterates, and let $r > -1 $ be a real number.   For $k\geq 0 $, the factorial power average 
\begin{equation}
\boxed{ \bar{x}_k = \frac{r+1}{{(k+1)}^{\overline{r+1}}}\sum_{i=0}^{k} {(i+1)}^{\overline{r}}x_i}\label{eq:factorial-power-average}
\end{equation}
is equal to the moving average
\begin{equation}\label{eq:factorial-power-average-online}
\bar{x}_{k+1}=\left(1-c_k\right)\bar{x}_{k}+c_k x_{k+1}, 
\end{equation}
 where  $c_{k}:=\displaystyle\frac{r+1}{k+r+1}.$
\end{prop}

\begin{proof}
We show by induction. For the base case, consider $k=0$. Then:
\begin{align*}
\bar{x}_{0} & =\left(1-c_{0}\right)\bar{x}_{-1}+c_{0}z_{0}\\
 & =\left(1-\frac{r+1}{r+1}\right)\bar{x}_{-1}+\frac{r+1}{r+1}z_{0} \;  = z_{0}.
\end{align*}
Likewise,
 we have that
\begin{align*}
\bar{x}_{0} & =\frac{r+1}{1^{\overline{r+1}}}\sum_{i=0}^{0}\left(i+1\right)^{\overline{r}}z_{i}
 \; =\frac{\left(r+1\right)1^{\overline{r}}}{1^{\overline{r+1}}}z_{0} \;  =z_{1},
\end{align*}
where we used the recursive property $\left(k+1\right)^{\overline{r}}=\left(k+r\right)\left(k+1\right)^{\overline{r-1}}$
to simplify.

For the inductive case, consider $k\geq1$ and suppose that $\bar{x}_{k-1}=\frac{r+1}{k^{\overline{r+1}}}\sum_{i=0}^{k-1}\left(i+1\right)^{\overline{r}}z_{i}.$
We may write the update as
\begin{align*}
\bar{x}_{k} & =\frac{r+1}{\left(k+1\right)^{\overline{r+1}}}\sum_{i=0}^{k}\left(i+1\right)^{\overline{r}}z_{i}.\\
 & =\frac{r+1}{\left(k+1\right)^{\overline{r+1}}}\sum_{i=1}^{k-1}\left(i+1\right)^{\overline{r}}z_{i}+\left(r+1\right)\frac{\left(k+1\right)^{\overline{r}}}{\left(k+1\right)^{\overline{r+1}}}z_{k}\\
 & =\frac{k^{\overline{r+1}}}{\left(k+1\right)^{\overline{r+1}}}\frac{r+1}{k^{\overline{r+1}}}\sum_{i=1}^{k-1}\left(i+1\right)^{\overline{r}}z_{i} \\
 & + \left(r+1\right)\frac{\left(k+1\right)^{\overline{r}}}{\left(k+1\right)^{\overline{r+1}}}z_{k}\\
 & =\frac{k^{\overline{r+1}}}{\left(k+1\right)^{\overline{r+1}}}\bar{x}_{k-1}+\left(r+1\right)\frac{\left(k+1\right)^{\overline{r}}}{\left(k+1\right)^{\overline{r+1}}}z_{k},
\end{align*}
where in the last line we used the induction hypothesis. To show the
equivalence to the moving average form $\bar{x}_{k}=\left(1-c_{k}\right)\bar{x}_{k-1}+c_{k}z_{k},$
we just need to show that
\[
c_{k}=\left(r+1\right)\frac{\left(k+1\right)^{\overline{r}}}{\left(k+1\right)^{\overline{r+1}}}\quad\mbox{{and} \ensuremath{\quad1-c_{k}} }=\frac{k^{\overline{r+1}}}{\left(k+1\right)^{\overline{r+1}}},
\]
 where $c_{k}=\frac{r+1}{k+r+1}$. 
By applying the recursive property~\eqref{eq:recurse_r} these two identities follow since
\begin{align*}
\left(r+1\right)\frac{\left(k+1\right)^{\overline{r}}}{\left(k+1\right)^{\overline{r+1}}} & =\left(r+1\right)\frac{\left(k+1\right)^{\overline{r}}}{\left(k+r+1\right)\left(k+1\right)^{\overline{r}}}\\
 & =c_{k}.
\end{align*}
For the second identity we use \eqref{eq:recurse_k} so that 
\begin{align*}
\frac{k^{\overline{r+1}}}{\left(k+1\right)^{\overline{r+1}}} & =\frac{k^{\overline{r+1}}}{\frac{k+r+1}{k}k^{\overline{r+1}}}
 \; =\frac{k}{k+r+1}\\
 & =1-\frac{r+1}{k+r+1} \;  =1-c_{k}.
\end{align*}
This concludes the inductive step and the proof.
\end{proof}
 \cite{shamirzhang2013} introduced the \emph{polynomial-decay averaging}~\eqref{eq:factorial-power-average-online}  for averaged SGD under the restriction that integer $r>0$. Proposition~\ref{prop:polydecayisfactorialpower} extends the result to non-integer values with a range of $r>-1$.  Next we use factorial power averaging to get state-of-the-art convergence results for \SGDM.

\subsection{Applying factorial powers}\vspace{-0.3em}
The any-time convergence of \SGDM  is a good case study for the application of the half-factorial powers.

\begin{thm}
\label{thm:SGD-nonsmooth-poly}
Let $f(x,\xi)$ be $G$-Lipschitz and convex in $x$. 
The projected \SGDM method~\eqref{eq:iam-main} with
   $\eta_{k}=\eta(k+1)^{\overline{-1/2}}$ for
 $\eta>0$  and
$c_{k+1}=1/(k+1)$ converges according to
\begin{align*}
\E \left[f(x_{n})-f(x_{*})\right]	\; \leq \; 
\frac{1}{2}\left(\eta^{-1}R^{2}+2\eta G^{2}\right)\left(n+2\right)^{\overline{-1/2}}.
\end{align*}
Furthermore, optimizing over $\eta$ gives 
$\eta=\sqrt{1/2}\frac{R}{G}$
and the resulting convergence 
\begin{align*}
\E \left[f(x_{n})-f(x_{*})\right] & \leq\sqrt{2}RG\left(n+2\right)^{\overline{-1/2}}.
\end{align*}
\end{thm}

This result is strictly tighter than the $\sqrt{2}RG/ \sqrt{n+1}$ convergence rate that arises from the use of square-root sequences~(see Theorem~\ref{thm:anytime-classic-sqrt} in the appendix) as used by \cite{primal_averaging}.
The use of half-factorial powers also yields more direct proofs, as inequalities are replaced with equalities in many places. For instance, when  $\eta_{k} = \eta / \sqrt{k+1}$, a bound of the following form arises in the proof:
\[
\sqrt{k+1}-\sqrt{k} \leq\frac{1}{2\sqrt{k}}.
\]
If factorial power step
sizes $\eta_{k}=\eta(k+1)^{\overline{-1/2}}$ are used instead, then this bounding
operation is replaced with an equality that we call the inverse difference property:
\[
\frac{1}{\left(k+1\right)^{\overline{-1/2}}}-\frac{1}{k^{\overline{-1/2}}}=\frac{1}{2}\frac{1}{k^{\overline{1/2}}}.
\]
The standard proof also requires summing the step sizes, requiring
another bounding operation
\[
\sum_{i=0}^{k}\frac{1}{\sqrt{i+1}}\leq2\sqrt{k+1}.
\]
Again when the factorial power step sizes are used instead, this inequality is replaced by the equality
$\sum_{i=0}^{k}(i+1)^{\overline{-1/2}}=2\left(k+1\right)^{\overline{1/2}}$. 

We can also use  factorial power momentum with $r=3$ to show that \SGDM converges at a rate of $\cO(1/n)$ for strongly-convex non-smooth problems in the following theorem. 
\begin{thm}
\label{thm:SGDM-non-smooth-str-cvx} 
Let $f(x,\xi)$ be $G$-Lipschitz and $\mu-$strongly convex in $x$ for every $\xi.$ The
projected \SGDM method~\eqref{eq:iam-main} with  $\eta_{k}=\frac{1}{\mu(k+1)}$ and $c_{k+1}=\frac{4}{k+4}$
satisfies
\[
\E\left[f(x_{n})-f(x_{*})\right]\leq \frac{2G^{2}}{\mu} \left(n+2\right)^{\overline{-1}} =  \frac{2G^{2}}{\mu(n+1)}.
\]
\end{thm}
This $\cO(1/n)$ rate of convergence is the fastest possible in this setting~\citep{Agarwal2009}. This rate of convergence has better constants than that established by using a different momentum scheme in \cite{primal_averaging}. Higher order averaging is also necessary to obtain this rate for the averaged SGD method, as established by~\citet{LacosteJulien2012} and~\cite{shamirzhang2013}, however in that case only $r=1$ averaging is necessary to obtain the same rate. 

\section{From Momentum to Acceleration}\vspace{-0.3em}
\label{sec:acceleration} A higher order $r$ for the factorial powers is useful when
the goal is to achieve convergence rates of the order $\cO(1/n^{r+1})$.
Methods using equal weighted $r=0$ momentum can not achieve convergence
rates faster than $\cO(1/n$), since that is the rate that they ``forget''
the initial conditions. To see this, note that in a sum $1/(n+1)\sum_{i=0}^{n}z_{i}$,
the $z_{0}$ value decays at a rate of $\cO(1/n)$. 
When using the order $r$ factorial power for averaging~\eqref{eq:factorial-power-average}, the initial conditions are forgotten at a rate of
 $\cO(1/n^{r+1})$. 
The need for $r=1$ averaging arises in a natural way when developing
accelerated optimization methods for non-strongly convex optimization,
where the best known rates are of the order $\cO(1/n^{2})$ obtained
by Nesterov's method. 
As with the \SGDM method, Nesterov's method can also be written in an
 equivalent iterate averaging form \citep{at2006}:
\begin{align}
y_{k} & =\left(1-c_{k+1}\right)x_{k}+c_{k+1}z_{k} \nonumber \\
z_{k+1} & =z_{k}- \rho_k \nabla f(y_{k})\nonumber \\
x_{k+1} & =\left(1-c_{k+1}\right)x_{k}+c_{k+1}z_{k+1}, \label{eq:Nesterov}
\end{align}
where $\rho_k$ are the step sizes, and initially $z_0 = x_0$.
In this formulation of Nesterov's method we can see that
the $x_{k}$ sequence uses iterate averaging of the  form~\eqref{eq:iam-main}.
To achieve accelerated rates with this method, the standard approach is to use  $\rho_k =1/(L{c_{k+1}})$
and to choose momentum constants $c_k$ that satisfy the inequality
\[ \textstyle
c_{k}^{-2}-c_{k}^{-1} \leq c_{k-1}^{-2}.
\]
This inequality is satisfied with equality when using the following
recursive formula:
\[ \textstyle
c_{k+1}^{-1}=\frac{1}{2} \left( 1+\sqrt{1+4c_{k-1}^{-1}}  \right),
\]
but the opaque nature and lack of closed form for this sequence is
unsatisfying. Remarkably, the sequence $c_{k+1}=2/(k+2)$ also satisfies
this inequality, as pointed out by \citet{tseng2008}, which is a simple application
of $r=1$ factorial power momentum. 
We show in the supplementary
material how using factorial powers together with the iterate averaging form of momentum gives a simple proof of convergence for this method, which uses the same proof technique and Lyapunov function as the proof of convergence of the regular momentum method \SGDM.
By leveraging the properties of factorial powers, 
the proof follows straightforwardly with no ``magic'' steps.

\begin{thm}\label{thm:nesterov-kplus1sqr}
Let $x_k$ be given by~\eqref{eq:Nesterov}. Let $f(x,\xi)$ be  $L$--smooth and convex. If we set $c_k = 2/(k+2)$ and $\rho_{k}=(k+1)/(2L)$ then
\begin{equation}
f(x_{n})-f(x_{*})	\leq\frac{2L}{n^{\overline{2}}}\left\Vert x_{0}-x_{*}\right\Vert ^{2}.
\end{equation}
\end{thm}
This matches the rate given by \citet{fista2009} asymptotically, and is faster than the rate given by Nesterov's estimate sequence approach \citet{Nesterov-convex} by a constant factor.

\section{Variance Reduction with Momentum}\vspace{-0.3em}
\begin{algorithm}
   \caption{Our proposed SVRGM method}
\label{alg:svrg+m}
\begin{algorithmic}
   \State {\bfseries Initialize:} $z_{m_{0-1}}^{0}=x_{m_{0}-1}^{0}=x_{0}$
   \For{$s=1,2,\ldots, $} \Comment{\texttt{outer-loop}}
   \State $\tilde{x}^{s-1}=x_{m_{s-1}-1}^{s-1}, \; x_{0}^{s}=x_{m_{s-1}-1}^{s-1},\;z_{0}^{s}=z_{m_{s-1}}^{s-1}$
   \State $\nabla f(\tilde{x}^{s-1})  =\frac{1}{n}\sum_{i=1}^{n}\nabla f_{i}(\tilde{x}^{s-1}).$ \Comment{\texttt{precompute}}
      \For{$t=0,1,\ldots, m_{s}-1$} \Comment{\texttt{inner-loop}}
      	 \State Sample $j$ uniformly at random 
      	 \State $g_{t}^{s}  =\nabla f_{j}(x_{t}^{s})-\left[\nabla f_{j}\left(\tilde{x}^{s-1}\right)-\nabla f(\tilde{x}^{s-1})\right] $
         \State $z_{t+1}^{s}  =z_{t}^{s}-\eta g_{t}^{s} $
         \State $x_{t+1}^{s}  =\left(1-c_{t+1}\right)x_{t}^{s}+c_{t+1}z_{t+1}$ \Comment{\texttt{Averaging}}
      \EndFor
   \EndFor
\end{algorithmic}
\end{algorithm}
Since factorial power momentum has clear advantages in situations where
averaging of the iterates is otherwise used, we further explore a
problem where averaging is necessary and significantly complicates
matters: the stochastic variance-reduced gradient method (SVRG).
The SVRG method \citep{svrg} is a double loop method, where the iterations in the inner loop resemble SGD steps, but with an additional additive variance reducing correction. In each outer loop, the average of the
iterates from the inner loop are used to form a new ``snapshot''
point. We propose the SVRGM method (Algorithm \ref{alg:svrg+m}).
This method modifies the improved SVRG++ formulation of \citet{SVRG++}
to further include the use of iterate averaging style momentum in the inner loop. See Algorithm~\ref{alg:svrg+m}.

Our formulation has a number of advantages over existing schemes. In
terms of simplicity, it includes no resetting operations\footnote{This is also a feature of the variant known as \emph{free-SVRG}~\cite{SVRGminib} }, so the $x$
and $z$ sequences start each outer loop at the values from the end
of the previous one. Additionally, the \emph{snapshot} $\tilde{x}$ is up-to-date, in the sense that it matches the final output point $x$ from the previous step, rather than being set to an average of points as in SVRG/SVRG++.
   
The non-strongly convex case is an application of non-integer factorial power
momentum. Using a large step size $\eta =1/6L$ we show in Theorem~\ref{thm:SVRGMconvex} That Algorithm~\ref{alg:svrg+m} converges at a favourable rate if we choose the momentum parameters $c_k$ corresponding to a 
$(k+1)^{\overline{1/2}}$ factorial power averaging of the iterates.
The strongly convex case in Theorem~\ref{thm:SVRGMstrconvex}
uses fixed momentum (i.e. an exponential moving average), since no rising factorial
sequence can give linear convergence rates. Both of these rates improve the constants non-trivially over the SVRG++ method.

\begin{thm}\label{thm:SVRGMconvex}
(non-strongly convex case) Let $f(x)=\frac{1}{n}\sum_{i=1}^{n}f_{i}(x)$
where each $f_{i}$ is $L$-smooth and convex. By setting $c_{t}=\frac{1/2+1}{t+1/2+1}$,
 $\eta=\frac{1}{6L}$,
and  $m_{s}=2m_{s-1}$ in Algorithm~\ref{alg:svrg+m} we have that
\[
\E\left[f(x_{m_{s}-1}^{S})-f^{*}\right]\leq\frac{f\left(x_{0}\right)-f^{*}}{2^{S}}+\frac{9L\left\Vert x_{0}-x_{*}\right\Vert ^{2}}{2^{S}m_{0}}
\]
\end{thm}
The non-strongly convex convergence rate is linear in the number of epochs, however each epoch is twice as long as the previous one, resulting in an overall $1/t$ rate.

\begin{thm} \label{thm:SVRGMstrconvex}
(strongly convex case) Let $\frac{1}{n}\sum_{i=1}^{n}f_{i}(x)$ where
each $f_{i}$ is $L$-smooth and $\mu$-strongly convex. Let $\kappa=L/\mu$.
By setting $m_{s}=6\kappa$, $c_{k}=\frac{5}{3}\frac{1}{4\kappa+1},$
and $\eta_{k}=1/(10L)$ in Algorithm~\ref{alg:svrg+m} we have that
\[
\E\left[f(\tilde{x}^{s})-f^{*}\right]\leq\left(\frac{3}{5}\right)^{S}\left[f(x_{0})-f(x_{*})+\frac{3}{4}\mu \, \delta_0\right],
\]
where $\delta_0 : =\left\Vert x_{0}-x_{*}\right\Vert ^{2}.$
\end{thm}

\section{Further Applications}\vspace{-0.3em}
\begin{figure*}
\includegraphics[width=1\textwidth]{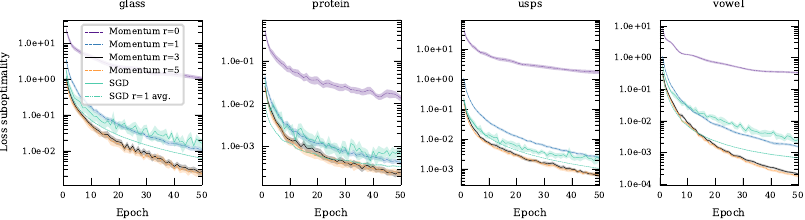}
\caption{\label{fig:sc-nonsmooth}Training loss sub-optimality on 4 LIBSVM
test problems, comparing SGD, SGD with $r=1$ post-hoc averaging to
SGD with factorial power momentum.}\vspace{-0.3em}
\end{figure*}
Factorial powers have applications across many areas of optimization theory. We detail two further instances of popular first order methods where factorial powers are particularly useful.

\subsection{Dual Averaging}
 Classical (non-stochastic) dual averaging uses updates of the form \citep{nes-da}:
\begin{align}
s_{k+1} &= s_{k}+\nabla f\left(x_{k}\right)\nonumber \\
x_{k+1}  &= \arg\min_{x}\left\{ \left\langle s_{k+1},x\right\rangle +\hat{\beta}_{k+1}\frac{\gamma}{2}\left\Vert x-x_{0}\right\Vert ^{2}\right\} ,\label{eq:dual-average}
\end{align}
where the sequence $\hat{\beta}_{k}$ is defined
recursively with $\hat{\beta}_{0}=\hat{\beta}_{1}=1$, and
$
\hat{\beta}_{k+1}=\hat{\beta}_{k+1}+1/\hat{\beta}_{k+1}.
$
This sequence grows approximately following the square root, as $\sqrt{2k-1}\leq\hat{\beta}_{k+1}\leq\frac{1}{1+\sqrt{3}}+\sqrt{2k-1}$
for $k\geq1$, and obeys a kind of summation property $\sum_{i=0}^{k}\frac{1}{\hat{\beta}_{i}}=\hat{\beta}_{k+1}$.
Nesterov's sequence has the disadvantage of not having a simple closed
form, but it otherwise provides tighter bounds than using $\beta_{k}=\sqrt{k+1}$.
In particular, the precise bound on the duality gap (as we show in Theorem~\ref{thm:dualaver} in the supplementary material) is given by
\begin{align*}
\max_{x,\left\Vert x\right\Vert \leq R}&\left\{\frac{1}{n+1}  \sum_{i=0}^{n}\left\langle \nabla f\left(x_{i}\right),x_{i}-x\right\rangle \right\}  \\
& \leq\left(\frac{\sqrt{2}}{\left(1+\sqrt{3}\right)}\frac{1}{(n+1)}+\frac{2}{\sqrt{n+1}}\right)RG.
\end{align*}
The factorial powers obey a similar summation relation, and they have the advantage of
an explicit closed form, which we exploit to give a strictly tighter convergence rate.
\begin{thm} \label{thm:dualaver-main}
After $n$ steps of the dual averaging method~\eqref{eq:dual-average} with $\hat{\beta}_{k}=1/\left(k+1\right)^{\overline{-1/2}}$ and $\gamma=G/R$ we have that
\begin{align*}
\max_{x,\left\Vert x\right\Vert \leq R}&\left\{ \frac{1}{n+1} \sum_{i=0}^{n}\left\langle \nabla f\left(x_{i}\right),x_{i}-x\right\rangle \right\}  \\
& \leq 2RG(n+2)^{\overline{-1/2}}<\frac{2RG}{\sqrt{n+1}}.
\end{align*}
\end{thm}
\subsection{Conditional Gradient Method}
Factorial power step size schemes have also arisen for the conditional gradient
method
\begin{align*}
p_{k+1} &= \arg\min_{p\in C}\left\langle p,\nabla f(x_{k})\right\rangle \\
 x_{k+1} & =\left(1-c_{k+1}\right)x_{k}+c_{k+1}p_{k+1}.
\end{align*}
For this method the most natural step sizes satisfy the following recurrence (``open loop'' step sizes)
$c_{k+1}~=~c_{k}-\frac{1}{2}c_{k}^{2},$
which \citet{cg-open-loop} note may be replaced with $c_{k+1}=1/(k+1)$.
Another approach that more closely approximates the open-loop steps is the factorial power weighting $c_{k+1}=2/(k+2)$ as used in \cite{pmlr-v28-jaggi13} and \cite{bach-cg}. 

\section{Experiments}\vspace{-0.3em}
For our experiments we compared the performance of factorial power momentum on a strongly-convex but non-smooth
machine learning problem: regularized multi-class support vector machines.
We consider two problems from the LIBSVM~\citep{chang2011libsvm} repository: \textit{PROTEIN}
and \textit{USPS}, and two from the UCI~\citep{Dua:2019} repository: \textit{GLASS} and \textit{VOWEL}.  We used batch-size 1 and the step sizes recommended by the theory
for both SGD with $r=1$ averaging, as well as SGD with factorial power momentum
as we developed in Theorem~\ref{thm:SGDM-non-smooth-str-cvx}. We induced strong convexity by using weight decay of strength $0.001$. The median as well as interquartile range bars from 40
runs are shown. Since our theory suggests $r=3$, we tested $r=0,1,3,5$
to verify that $r=3$ is the best choice. The results are shown in
Figure \ref{fig:sc-nonsmooth}. We see that when using factorial power momentum,
using $r=0,1$ is worse than $r=3$, and using $r=5$ is no better
that $r=3$, so the results agree with our theory. The momentum method
also performs a little better than SGD with post-hoc averaging, however
it does appear to be substantially more variable between runs, as the interquartile range shows.
We provide further experiments covering the SVRGM method in the supplementary material.

\section{Conclusion}\vspace{-0.3em}
Factorial powers are a flexible and broadly applicable tool for establishing tight convergence rates as well as simplifying proofs. As we have shown, they have broad applicability both for stochastic optimization and beyond.

\bibliographystyle{plainnat}
\bibliography{../iteration_weighted}

\newpage
\appendix

\onecolumn
\section{Proof of Properties of Factorial Powers}
\label{sec:properties}

We recall that we define the factorial powers using the gamma function
\begin{equation}\label{eq:factgammarecall}
k^{\overline{r}} \; := \; \frac{\Gamma(k+r)}{\Gamma(k)}, \quad \mbox{where  } \Gamma(k)\; := \; \int_{0}^{-\infty} x^{k-1} e^{-x} dx, \quad \mbox{for } k +r > 0\mbox{ and } k \geq 1.
\end{equation}
We also extend the definition and set
$0^{\overline{r}}=0$ except for $0^{\overline{0}}=1$.
 We restrict $k +r >0$ and $k \geq 1$ in~\eqref{eq:factgammarecall} because the gamma function $\Gamma(z)$ is only well defined for $z>0$.

We will use the following well known property of the gamma function 
\begin{equation} \label{eq:gammarec}
\Gamma(k+1) \; = \; k\Gamma(k),
\end{equation}
that follows by integration by parts.

We now give the proof of all the properties in Table~\ref{tab:prop}.

\begin{prop}
For $k\geq 1$ and $k+r >0$ we have that following
recursive properties:
\begin{equation}
\left(k+1\right)^{\overline{r}}=\frac{k+r}{k}k^{\overline{r}},\label{eq:recurse_k-1}
\end{equation}
\begin{equation}
\left(k+1\right)^{\overline{r}}=\left(k+r\right)\left(k+1\right)^{\overline{r-1}}.\label{eq:recurse_r-1}
\end{equation}
\end{prop}

\begin{proof}
Using the definition directly

\begin{eqnarray*}
\left(k+1\right)^{\overline{r}} & = &\frac{\Gamma(k+r+1)}{\Gamma(k+1)}\\
 & \overset{\eqref{eq:gammarec}}{=} &\frac{\Gamma(k+r)\left(k+r\right)}{\Gamma(k)k}\\
 & = &\frac{k+r}{k}k^{\overline{r}},
\end{eqnarray*}
and
\[
\left(k+1\right)^{\overline{r}}=\frac{\Gamma(k+1+r)}{\Gamma(k+1)}\overset{\eqref{eq:gammarec}}{=}\frac{\Gamma(k+r)(k+r)}{\Gamma(k+1)}=\left(k+r\right)\left(k+1\right)^{\overline{r-1}}.
\]
\end{proof}
\begin{prop}
For $k\geq 1$ and $k+r >0$ we have that following difference property
\end{prop}

\begin{equation}
\left(k+1\right)^{\overline{r}}-k^{\overline{r}}=r\left(k+1\right)^{\overline{r-1}}. \label{eq:diffprop-1}
\end{equation}

\begin{proof}
We apply the recursive property in $k$, then in $r$

\begin{align*}
\left(k+1\right)^{\overline{r}}-k^{\overline{r}} & =\frac{k+r}{k}k^{\overline{r}}-k^{\overline{r}}\\
 & =\frac{r}{k}k^{\overline{r}}\\
 & =\frac{1}{\frac{k+r}{k}}\frac{r}{k}\left(k+1\right)^{\overline{r}}\\
 & =r\frac{1}{k+r}\left(k+1\right)^{\overline{r}}\\
 & =r\left(k+1\right)^{\overline{r-1}}.
\end{align*}
\end{proof}
\begin{prop}
\label{prop:ratio_property}
For $k\geq 1$, $k+r >0$ and $k+r+q >0$ we have that following
ratio property
\[
\frac{k^{\overline{r+q}}}{k^{\overline{r}}}=\left(k+r\right)^{\overline{q}},
\]
\end{prop}

\begin{proof}
\begin{align*}
\frac{k^{\overline{r+q}}}{k^{\overline{r}}} & =\frac{\frac{\Gamma(k+r+q)}{\Gamma(k)}}{\frac{\Gamma(k+r)}{\Gamma(k)}}\\
 & =\frac{\Gamma(k+r+q)}{\Gamma(k+r)}\\
 & =\left(k+r\right)^{\overline{q}}.
\end{align*}
\end{proof}
\begin{prop}
For integers $b\geq a\geq 1$ such that $a+r>0$ we have the following
summation property
\[
\sum_{i=a}^{b}i^{\overline{r}}=\frac{1}{r+1}b^{\overline{r+1}}-\frac{1}{r+1}\left(a-1\right)^{\overline{r+1}}.
\]
\end{prop}

\begin{proof}
This property is a direct consequence of telescoping the difference property:
\begin{align*}
\left(r+1\right)\sum_{i=a}^{b}i^{\overline{r}}= & \sum_{i=a-1}^{b-1}(r+1)(i+1)^{\overline{r}}\\
= & \sum_{i=a-1}^{b-1}\left[\left(i+1\right)^{\overline{r+1}}-i^{\overline{r+1}}\right]\\
= & \left(b-1+1\right)^{\overline{r+1}}-\left(a-1\right)^{\overline{r+1}}\\
= & b^{\overline{r+1}}-\left(a-1\right)^{\overline{r+1}}
\end{align*}
\end{proof}
\begin{prop}
For $k \geq 1$ we have that following inverse difference property
\begin{equation}
\frac{1}{\left(k+1\right)^{\overline{-1/2}}}-\frac{1}{k^{\overline{-1/2}}}=\frac{1}{2}\frac{1}{k^{\overline{1/2}}}.
\end{equation}
\end{prop}

\begin{proof}
We apply the inverse property followed by the difference property
then the inverse property again:
\begin{align*}
\frac{1}{\left(k+1\right)^{\overline{-1/2}}}-\frac{1}{k^{\overline{-1/2}}} & =\left(k+1-\frac{1}{2}\right)^{\overline{1/2}}-\left(k-\frac{1}{2}\right)^{\overline{1/2}}\\
 & =\frac{1}{2}\left(k+1/2\right)^{\overline{-1/2}}\\
 & =\frac{1}{2}\frac{1}{k^{\overline{1/2}}}.
\end{align*}
\end{proof}

\begin{lem}
\label{lem:ck_iterateweighting} Let $k \geq 1$, $r\geq0$ and $j\geq0$. 
Consider the sequence
\[
c_{k}=\frac{r+1}{k+j+r}.
\]
It follows that
\[
\frac{1-c_{k}}{c_{k}}(k+j)^{\overline{r}}=\frac{1}{c_{k-1}}(k+j-1)^{\overline{r}}.
\]
\end{lem}

\begin{proof}
Simplifying:
\begin{align*}
\left(\frac{1}{c_{k}}-1\right)(k+j)^{\overline{r}} & =\left(\frac{k+j+r}{r+1}-1\right)(k+j)^{\overline{r}}\\
 & =\left(\frac{k+j+r-r-1}{r+1}\right)(k+j)^{\overline{r}}\\
 & =\frac{k+j-1}{r+1}(k+j)^{\overline{r}}\\
 & =\frac{k+j+r-1}{r+1}\frac{k+j-1}{k+j+r-1}(k+j)^{\overline{r}}\\
 & =\frac{1}{c_{k-1}}\frac{k+j-1}{k+j+r-1}(k+j)^{\overline{r}}.
\end{align*}
Now applying the recursion property Eq. (\ref{eq:recurse_k}) gives:
\[
\frac{k+j-1}{k+j+r-1}(k+j)^{\overline{r}}=(k+j-1)^{\overline{r}},
\]
giving the result.
\end{proof}

\section{Convergence Theorems for the Projected \SGDM}
\begin{thm}
\label{thm:momentum}Consider the projected \SGDM method
\begin{align}
x_{k} & =\left(1-c_{k}\right)x_{k-1}+c_{k}z_{k},\label{eq:online-average-ap}\\
z_{k+1} & =\varPi_{C}\left(z_{k}-\eta_{k}\nabla f(x_{k},\xi_{k})\right),\nonumber 
\end{align}
where $0 < c_{k}\leq 1.$ If each $f(\cdot,\xi)$ is convex and $G$-Lipschitz
then
\begin{align*}
\left\Vert z_{k+1}-x_{*}\right\Vert ^{2} & \leq\left\Vert z_{k}-x_{*}\right\Vert ^{2}+\eta_{k}^{2}G^{2}\\
 & -2\frac{1}{c_{k}}\eta_{k}\left[f(x_{k},\xi_{k})-f(x_{*},\xi_{k})\right]+2\left(\frac{1}{c_{k}}-1\right)\eta_{k}\left[f(x_{k-1},\xi_{k})-f(x_{*},\xi_{k})\right].
\end{align*}
\end{thm}

\begin{proof}
We start with $z_{k+1}$ instead of the usual expansion in terms of
$x_{k+1}$:
\begin{align*}
\left\Vert z_{k+1}-x_{*}\right\Vert ^{2} & =\left\Vert \varPi_{C}\left(z_{k}-\eta_{k}\nabla f(x_{k},\xi_{k})\right)-\varPi_{C}\left(x_{*}\right)\right\Vert ^{2}\\
 & \leq\left\Vert z_{k}-\eta_{k}\nabla f(x_{k},\xi_{k})-x_{*}\right\Vert ^{2}\\
 & =\left\Vert z_{k}-x_{*}\right\Vert ^{2}-2\eta_{k}\left\langle \nabla f(x_{k},\xi_{k}),z_{k}-x_{*}\right\rangle +\eta_{k}^{2}G^{2}\\
 & =\left\Vert z_{k}-x_{*}\right\Vert ^{2}-2\eta_{k}\left\langle \nabla f(x_{k},\xi_{k}),x_{k}-\left(\frac{1}{c_{k}}-1\right)\left(x_{k-1}-x_{k}\right)-x_{*}\right\rangle +\eta_{k}^{2}G^{2}\\
 & =\left\Vert z_{k}-x_{*}\right\Vert ^{2}+\eta_{k}^{2}G^{2}\\
 & -2\eta_{k}\left\langle \nabla f(x_{k},\xi_{k}),x_{k}-x_{*}\right\rangle -2\eta_{k}\left(\frac{1}{c_{k}}-1\right)\left\langle \nabla f(x_{k},\xi_{k}),x_{k}-x_{k-1}\right\rangle 
\end{align*}
Using the following two convexity inequalities
\[
\left\langle \nabla f\left(x_{k},\xi_{k}\right),x_{*}-x_{k}\right\rangle \leq f(x_{*},\xi_{k})-f(x_{k},\xi_{k})
\]
\[
\left\langle \nabla f\left(x_{k},\xi_{k}\right),x_{k-1}-x_{k}\right\rangle \leq f(x_{k-1},\xi_{k})-f(x_{k},\xi_{k})
\]
combined with $(1/c_{k}-1)\geq0$ gives 
\begin{align*}
\left\Vert z_{k+1}-x_{*}\right\Vert ^{2} & \leq\left\Vert z_{k}-x_{*}\right\Vert ^{2}+\eta_{k}^{2}G^{2}\\
 & -2\eta_{k}\left[f(x_{k},\xi_{k})-f(x_{*},\xi_{k})\right]-2\left(\frac{1}{c_{k}}-1\right)\eta_{k}\left[f(x_{k},\xi_{k})-f(x_{k-1},\xi_{k})\right].
\end{align*}
Now rearranging further gives the result
\begin{align*}
\left\Vert z_{k+1}-x_{*}\right\Vert ^{2} & \leq\left\Vert z_{k}-x_{*}\right\Vert ^{2}+\eta_{k}^{2}G^{2}\\
 & -2\frac{1}{c_{k}}\eta_{k}\left[f(x_{k},\xi_{k})-f(x_{*},\xi_{k})\right]+2\left(\frac{1}{c_{k}}-1\right)\eta_{k}\left[f(x_{k-1},\xi_{k})-f(x_{*},\xi_{k})\right].
\end{align*}
\end{proof}
\begin{cor}
Consider the Lyapunov function:
\[
A_{k}=\left\Vert z_{k}-x_{*}\right\Vert ^{2}+\frac{2}{c_{k-1}}\eta_{k-1}\left[f(x_{k-1})-f(x_{*})\right]
\]
 If for $k\geq2,$ 
\begin{equation}
\left(\frac{1}{c_{k}}-1\right)\eta_{k}\leq\frac{1}{c_{k-1}}\eta_{k-1},\label{eq:momentum_requirement}
\end{equation}
and for $k=1$ we have $\left(\frac{1}{c_{1}}-1\right)\eta_{1}\leq0$,
then \SGDM steps statisfy the following relation for $k\geq1$.

\begin{align*}
\E_{\xi_{k}}\left[A_{k+1}\right] & \leq A_{k}+\eta_{k}^{2}G^{2},
\end{align*}
when each $f(\cdot,\xi)$ is convex and $G$-Lipschitz.
\end{cor}

\begin{cor}
Let $\E[\cdot]$ denote the expectation with respect to all $\xi_{i}$,
with $i\leq n$. Suppose that the constraint set $C$ is contained
in an $R$-ball around the origin. Then telescoping and applying the
law of total expectation gives:
\begin{align}
\E\left\Vert z_{n+1}-x_{*}\right\Vert ^{2}+\frac{2}{c_{n}}\eta_{n}\E\left[f(x_{n})-f(x_{*})\right] & \leq R^{2}+\sum_{i=0}^{n}\eta_{i}^{2}G^{2}\label{eq:telescoping-sgdm}
\end{align}
\end{cor}

\subsection{Proof of Theorem~\ref{thm:SGD-nonsmooth-poly}: Any-time convergence
with factorial power step sizes}

\begin{thm}
Consider the projected \SGDM method Eq. \ref{eq:iam-main}.
When $\eta_{k}=\sqrt{1/2}\frac{R}{G}(k+1)^{\overline{-1/2}}$ and
$c_{k}=1/(k+1)$, when each $f(x,\xi)$ is $G$-Lipschitz, convex and the
constraint set $C$ is contained within an $R$-ball around $x_{0}$, then:
\begin{align*}
\E \left[f(x_{n})-f(x_{*})\right] & \leq\sqrt{2}RG\left(n+2\right)^{\overline{-1/2}}\leq\frac{\sqrt{2}RG}{\sqrt{n+1}}.
\end{align*}
\end{thm}

\begin{proof}
Consider Theorem \ref{thm:momentum} in expectation conditioned on
$\xi_{k}$:
\begin{align*}
\E \left\Vert z_{k+1}-x_{*}\right\Vert ^{2} & \leq\left\Vert z_{k}-x_{*}\right\Vert ^{2}+\eta_{k}^{2}G^{2}\\
 & -2\frac{1}{c_{k}}\eta_{k}(f(x_{k})-f(x_{*}))+2\left(\frac{1}{c_{k}}-1\right)\eta_{k}(f(x_{k-1})-f(x_{*})).
\end{align*}
We will use a step size $\eta_{k}=\eta(k+1)^{\overline{-1/2}}$ for
some constant $\eta$, and multiply this expression by $1/(k+1)^{\overline{-1/2}}$:

\begin{align}
\frac{1}{(k+1)^{\overline{-1/2}}} \E \left\Vert z_{k+1}-x_{*}\right\Vert ^{2} & \leq\frac{1}{(k+1)^{\overline{-1/2}}}\left\Vert z_{k}-x_{*}\right\Vert ^{2}+(k+1)^{\overline{-1/2}}\eta^{2}G^{2}\nonumber \\
 & -2\frac{1}{c_{k}}\eta\left(f(x_{k})-f(x_{*})\right)+2\left(\frac{1}{c_{k}}-1\right)\eta\left(f(x_{k-1})-f(x_{*})\right).\label{eq:induction-poly}
\end{align}
Now we prove the result by induction. First consider the base case
$k=0$. 
Since $1^{\overline{-1/2}} = \frac{\Gamma(1/2)}{\Gamma(1)} = \sqrt{\pi}$ which follows since $\Gamma(1/2) = \sqrt{\pi}$ and $\Gamma(1) = 1$ we have that
\[
\frac{1}{1^{-1/2}}\left\Vert z_{0}-z_{*}\right\Vert ^{2}=\frac{1}{\sqrt{\pi}}\left\Vert z_{0}-z_{*}\right\Vert ^{2}\leq\frac{1}{\sqrt{\pi}}R^{2}.
\]
Consequently taking $k=0$ in (\ref{eq:induction-poly}) gives
\begin{equation}
\begin{aligned}\frac{1}{1^{\overline{-1/2}}} \E \left\Vert z_{1}-x_{*}\right\Vert ^{2} & \leq\frac{1}{1^{\overline{-1/2}}}\left\Vert z_{1}-z_{*}\right\Vert ^{2}+(1)^{\overline{-1/2}}\eta^{2}G^{2}-2\frac{1}{c_{0}}\eta\left(f(x_{0})-f(x_{*})\right)\\
 & + 2\left(\frac{1}{c_{0}}-1\right)\eta\left(f(x_{0-1})-f(x_{*})\right) \\
 & \leq\frac{1}{\sqrt{\pi}}R+1{}^{\overline{-1/2}}\eta^{2}G^{2}-2\eta\left(f(x_{0})-f(x_{*})\right).
\end{aligned}
\label{eq:basecase-poly}
\end{equation}
Inductive case: consider the case $k\geq1$. To facilitate telescoping
we want $\frac{1}{k^{\overline{-1/2}}}\left\Vert z_{k}-z_{*}\right\Vert ^{2}$
on the right, so to this end we rewrite
\begin{align}
\frac{1}{(k+1)^{\overline{-1/2}}}\left\Vert z_{k}-z_{*}\right\Vert ^{2} & =\frac{1}{k^{\overline{-1/2}}}\left\Vert z_{k}-z_{*}\right\Vert ^{2}+\left(\frac{1}{(k+1)^{\overline{-1/2}}}-\frac{1}{k^{\overline{-1/2}}}\right)\left\Vert z_{k}-z_{*}\right\Vert ^{2} \nonumber \\
 & \leq\frac{1}{k^{\overline{-1/2}}}\left\Vert z_{k}-z_{*}\right\Vert ^{2}+\left(\frac{1}{(k+1)^{\overline{-1/2}}}-\frac{1}{k^{\overline{-1/2}}}\right)R^{2}.\label{eq:aodjaaa4}
\end{align}
Now since $k\geq1$ we can apply the inverse difference property
\[
\frac{1}{\left(k+1\right)^{\overline{-1/2}}}-\frac{1}{k^{\overline{-1/2}}}=\frac{1}{2}\frac{1}{k^{\overline{1/2}}}
\]
which when used with~\eqref{eq:aodjaaa4} and then inserting the result in~\eqref{eq:induction-poly} gives
\begin{align*}
\frac{1}{\left(k+1\right)^{\overline{-1/2}}} \E \left\Vert z_{k+1}-x_{*}\right\Vert ^{2} & \leq\frac{1}{k^{\overline{-1/2}}}\left\Vert z_{k}-z_{*}\right\Vert ^{2}+\frac{1}{2}\frac{1}{k^{\overline{1/2}}}R^{2}+(k+1)^{\overline{-1/2}}\eta^{2}G^{2}\\
 & -2\frac{1}{c_{k}}\eta\left(f(x_{k})-f(x_{*})\right)+2\left(\frac{1}{c_{k}}-1\right)\eta\left(f(x_{k-1})-f(x_{*})\right).
\end{align*}
Since $c_{k}=1/(k+1)$ and $\frac{1}{k^{\overline{1/2}}}=\left(k+\frac{1}{2}\right)^{\overline{-1/2}}$
we have that
\begin{align}
\frac{1}{\left(k+1\right)^{\overline{-1/2}}} \E \left\Vert z_{k+1}-x_{*}\right\Vert ^{2} & \leq\frac{1}{k^{\overline{-1/2}}}\left\Vert z_{k}-z_{*}\right\Vert ^{2}+\frac{1}{2}\left(k+\frac{1}{2}\right)^{\overline{-1/2}}R^{2}+(k+1)^{\overline{-1/2}}\eta^{2}G^{2}\nonumber \\
 & -2(k+1)\eta\left(f(x_{k})-f(x_{*})\right)+2k\eta\left(f(x_{k-1})-f(x_{*})\right).\label{eq:induction-k-step-poly}
\end{align}
Now taking expectation and adding up both sides of (\ref{eq:induction-k-step-poly})
from $1$ to $n$ and using telescopic cancellation gives
\[
\begin{aligned}
\frac{1}{\left(n+1\right)^{\overline{-1/2}}} \E \left\Vert z_{n+1}-x_{*}\right\Vert ^{2} & \leq\frac{1}{1^{\overline{-1/2}}}\E \left\Vert z_{1}-z_{*}\right\Vert ^{2} \\
& +\frac{1}{2}R^{2}\sum_{i=1}^{n}\left(i+\frac{1}{2}\right)^{\overline{-1/2}}+\sum_{i=1}^{n}(i+1)^{\overline{-1/2}}\eta^{2}G^{2}\\
 & + 2\eta\left(f(x_{0})-f(x_{*})\right)-2(n+1)\eta \E[f(x_{n})-f(x_{*})].
\end{aligned}
\]
Now using the base case (\ref{eq:basecase-poly}) we have that
\begin{equation}
\begin{aligned}
 & \frac{1}{\left(n+1\right)^{\overline{-1/2}}} \E \left\Vert z_{n+1}-x_{*}\right\Vert ^{2} \\
 & \leq R^{2}+1^{\overline{-1/2}}\eta^{2}G^{2}-2\eta\left(f(x_{0})-f(x_{*})\right)-2(n+1)\eta \E[f(x_{n})-f(x_{*})] \\
 & \quad+\frac{1}{2}R^{2}\sum_{i=1}^{n}\left(i+\frac{1}{2}\right)^{\overline{-1/2}}+\sum_{i=1}^{n}(i+1)^{\overline{-1/2}}\eta^{2}G^{2}+2\eta\left(f(x_{0})-f(x_{*})\right)\\
 & =\frac{1}{\sqrt{\pi}}R^{2}+\frac{1}{2}R^{2}\sum_{i=1}^{n}\left(i+\frac{1}{2}\right)^{\overline{-1/2}}+\sum_{i=0}^{n}(i+1)^{\overline{-1/2}}\eta^{2}G^{2}-2(n+1)\eta \E[f(x_{n})-f(x_{*})].
\end{aligned}
\label{eq:inductiong-poly-finalstep}
\end{equation}
Using the summation property Eq. (\ref{eq:general_sum}) we have that
\[
\begin{aligned}\sum_{i=1}^{n}\left(i+1/2\right)^{\overline{-1/2}} & =2\left(n+1/2\right)^{\overline{1/2}}-2\left(3/2\right)^{\overline{1/2}}\\
 & =2\left(n+1/2\right)^{\overline{1/2}}-\frac{4}{\sqrt{\pi}}\\
 & \leq2\left(n+1\right)^{\overline{1/2}}-\frac{4}{\sqrt{\pi}}
\end{aligned}
\]
and furthermore
\[
\sum_{i=0}^{n}(i+1)^{\overline{-1/2}} \; =\;
\sum_{i=1}^{n+1}i^{\overline{-1/2}} \; \leq \;  2\left(n+1\right)^{\overline{1/2}}.
\]
So after dividing by $2(n+1)\eta$: 
\begin{align*}
\E\left[f(x_{n})-f(x_{*})\right] & \leq\frac{1}{2}\left(\frac{1}{\eta}R^{2}+2\eta G^{2}\right)\frac{\left(n+1\right)^{\overline{1/2}}}{n+1}
\end{align*}
We now use the ratio property on:
\[
\frac{\left(n+1\right)^{\overline{1/2}}}{n+1}=\frac{\left(n+1\right)^{\overline{1-1/2}}}{\left(n+1\right)^{\overline{1}}}=\left(n+2\right)^{\overline{-1/2}},
\]
 and solve for the best step size $\eta$, which is $\eta=\sqrt{1/2}\frac{R}{G}$
giving: 
\begin{align*}
\E \left[f(x_{n})-f(x_{*})\right] & \leq\sqrt{2}RG\left(n+2\right)^{\overline{-1/2}}\\
 & < \frac{\sqrt{2}RG}{\sqrt{n+1}}.
\end{align*}
\end{proof}

\subsection{Any-time convergence with standard step sizes:}

\begin{thm}\label{thm:anytime-classic-sqrt}
Let $f(x,\xi)$ be $G$-Lipschitz and convex for every $\xi.$
When $\eta_{k}=\frac{R}{G\sqrt{2(k+1)}}$ and $c_{k}=\frac{{1}}{k+1}$
in the projected \SGDM method~\eqref{eq:iam-main} we have that
\begin{align}\label{eq:anytime-classic-sqrt}
\E\left[f(x_{n})-f(x_{*})\right] & \leq\frac{\sqrt{2}RG}{\sqrt{n+1}}.
\end{align}
\end{thm}

\begin{proof}
We use $\eta_{k}=\eta/\sqrt{k+1}$ and $c_{k}=\frac{{1}}{k+1}$ in
the result from Theorem \ref{thm:momentum}, taking expectation and
multiplying both sides by $\sqrt{{k+1}}$ gives

\begin{align}
\sqrt{k+1}\E\left\Vert z_{k+1}-x_{*}\right\Vert ^{2} & \leq\sqrt{k+1}\left\Vert z_{k}-x_{*}\right\Vert ^{2}+\frac{1}{\sqrt{k+1}}\eta^{2}G^{2}\nonumber \\
 & -2(k+1)\eta \E\left[f(x_{k})-f(x_{*})\right]+2k\eta \E\left[f(x_{k-1})-f(x_{*})\right].\label{eq:induct-classic-sqrt}
\end{align}

For $k=0$ the above gives
\begin{align}
\E\left\Vert z_{1}-x_{*}\right\Vert ^{2} & \leq R^{2}+\eta^{2}G^{2} -2\eta \E\left[f(x_{0})-f(x_{*})\right].\label{eq:base-classic-sqrt}
\end{align}

For $k\geq1$, from concavity of the square root function
\begin{equation}\label{eq:sqrconcave}
\sqrt{k+1}-\sqrt{k}\leq\frac{1}{2\sqrt{k}},
\end{equation} 
we have that
\[
\sqrt{k+1}\left\Vert z_{k}-x_{*}\right\Vert ^{2}\leq\left(\sqrt{{k}}+\frac{1}{2\sqrt{k}}\right)\left\Vert z_{k}-x_{*}\right\Vert ^{2}\leq\sqrt{{k}}\left\Vert z_{k}-x_{*}\right\Vert ^{2}+\frac{1}{2\sqrt{k}}R^{2}.
\]
Plugging the above into~\eqref{eq:induct-classic-sqrt} gives  
\begin{align*}
\sqrt{k+1}\E\left\Vert z_{k+1}-x_{*}\right\Vert ^{2} & \leq\sqrt{k}\left\Vert z_{k}-x_{*}\right\Vert ^{2}+\left(\frac{1}{2\sqrt{k}}\right)R^{2}+\frac{1}{\sqrt{k+1}}\eta^{2}G^{2}\\
 & -2(k+1)\eta \E\left[f(x_{k})-f(x_{*})\right]+2k\eta \E\left[f(x_{k-1})-f(x_{*})\right].
\end{align*}
Now we telescope for $1$ to $n$ giving: 
\begin{align*}
\sqrt{n+1}\E\left\Vert z_{n+1}-x_{*}\right\Vert ^{2} & \leq\left\Vert z_{1}-x_{*}\right\Vert ^{2}+\sum_{i=1}^n\left(\frac{1}{2\sqrt{i}}\right)R^{2}+\sum_{i=1}^n\frac{1}{\sqrt{i+1}}\eta^{2}G^{2}\\
 & -2(n+1)\eta \E\left[f(x_{n})-f(x_{*})\right]+2\eta \E\left[f(x_{0})-f(x_{*})\right].
\end{align*}
Using the base case~\eqref{eq:base-classic-sqrt} we have that
\begin{align*}
\sqrt{n+1}\E\left\Vert z_{n+1}-x_{*}\right\Vert ^{2} & \leq R^{2}+\sum_{i=1}^n\left(\frac{1}{2\sqrt{i}}\right)R^{2}+\sum_{i=0}^n\frac{1}{\sqrt{i+1}}\eta^{2}G^{2}\\
 & -2(n+1)\eta \E\left[f(x_{n})-f(x_{*})\right].
\end{align*}
Now using the integral bounds
\begin{align*}
\sum_{i=1}^{n}\frac{1}{\sqrt{i}}&\leq 2(\sqrt{n}-1),\\
\sum_{i=0}^{n}\frac{1}{\sqrt{i+1}}& \leq2\sqrt{n+1},
\end{align*}
and re-arranging gives
\begin{align*}
2(n+1)\eta \E\left[f(x_{n})-f(x_{*})\right]  & \leq \sqrt{n}R^{2}+2\sqrt{n+1}\eta^{2}G^{2} -\sqrt{n+1}\E\left\Vert z_{n+1}-x_{*}\right\Vert ^{2} \\
&\leq \sqrt{n}R^{2}+2\sqrt{n+1}\eta^{2}G^{2} .
\end{align*}
Dividing through by $2(n+1)\eta$ gives
\begin{align*}
2(n+1)\eta \E\left[f(x_{n})-f(x_{*})\right]  
&\leq \frac{\sqrt{n}}{2(n+1)\eta}R^{2}+\frac{1}{\sqrt{n+1}}\eta G^{2} \\
&\leq \frac{1}{\sqrt{n+1}}\left( \frac{1}{2\eta}R^{2}+\eta G^{2}\right).
\end{align*}
Minimizing the above in $\eta$ gives $\eta =R/(\sqrt{2} G)$ which gives~\eqref{eq:anytime-classic-sqrt} and concludes the proof.
\end{proof}

\section{Strongly Convex Convergence}

Consider again the \SGDM method with a projection step given by
\begin{align*}
z_{k+1} & =\Pi_{C}\left(z_{k}-\eta_{k}\nabla f(x_{k},\xi_{k})\right),\\
x_{k+1} & =\left(1-c_{k+1}\right)x_{k}+c_{k+1}z_{k+1}.
\end{align*}

\begin{lem}
\label{lem:lyapunov_relation}For $\lambda_{k+1}=\frac{k+2}{2}$ and
$c_{k+1}=\frac{4}{k+4}$ we have that
\[
A_{k+1}:=\left\Vert x_{k+1}-x_{*}+\lambda_{k+1}\left(x_{k+1}-x_{k}\right)\right\Vert ^{2}=\left\Vert 2z_{k+1}-x_{k}-x_{*}\right\Vert ^{2}
\]
\end{lem}

\begin{proof}
The relation follows from substitution of the known relations:
\begin{align*}
A_{k+1} & =\left\Vert x_{k+1}-x_{*}+\lambda_{k+1}\left(x_{k+1}-x_{k}\right)\right\Vert ^{2}\\
 & =\left\Vert \left(\lambda_{k+1}+1\right)x_{k+1}-\lambda_{k+1}x_{k}-x_{*}\right\Vert ^{2}\\
 & =\left\Vert \left(\lambda_{k+1}+1\right)\left(\left(1-c_{k+1}\right)x_{k}+c_{k+1}z_{k+1}\right)-\lambda_{k+1}x_{k}-x_{*}\right\Vert ^{2}\\
 & =\left\Vert \left(\lambda_{k+1}+1\right)\left(\left(1-c_{k+1}\right)x_{k}+c_{k+1}z_{k+1}\right)+\left[\left(\lambda_{k+1}+1\right)\left(1-c_{k+1}\right)-\lambda_{k+1}\right]x_{k}-x_{*}\right\Vert ^{2}\\
 & =\left\Vert \left(\lambda_{k+1}+1\right)c_{k+1} z_{k+1} +\left[\left(\lambda_{k+1}-\lambda_{k+1}c_{k+1}+1-c_{k+1}\right)x_{k}-\lambda_{k+1}x_{k}\right]-x_{*}\right\Vert ^{2}\\
 & =\left\Vert \left(\lambda_{k+1}+1\right)c_{k+1}z_{k+1}+\left[\left(1-\left(\lambda_{k+1}+1\right)c_{k+1}\right)x_{k}\right]-x_{*}\right\Vert ^{2}.
\end{align*}
Now using
\[
\left(\lambda_{k+1}+1\right)c_{k+1}=\left(\frac{k+2}{2}+1\right)\frac{4}{k+4}=\frac{k+4}{2}\frac{4}{k+4}=2,
\]
gives
\begin{align*}
A_{k+1} = \left\Vert 2z_{k+1}-x_{k}-x_{*}\right\Vert ^{2}.
\end{align*}
\end{proof}

\subsection{Proof of Theorem~\ref{thm:SGDM-non-smooth-str-cvx}}
\begin{thm}
Let $f(x,\xi)$ be $G$-Lipschitz and $\mu-$strongly convex in $x$ for every $\xi.$ The
projected \SGDM method~\eqref{eq:iam-main} with  $\eta_{k}=\frac{1}{\mu(k+1)}$ and $c_{k+1}=\frac{4}{k+4}$
satisfies
\[
E\left[f(x_{n})-f(x_{*})\right]\leq\frac{2G^{2}}{\mu(n+1)}.
\]
\end{thm}

\begin{proof}
We will define a few constants to reduce notational clutter. Let
\[
\rho_{k}=\frac{k-1}{k+1},\,\text{and }\lambda_{k+1}=\frac{k+2}{2}.
\]
We will first apply the contraction property of the projection operator
(using the fact that $x_{k}$ and $x_{*}$ are always within the constraint
set) so that
\begin{align*}
A_{k+1} & =\left\Vert 2z_{k+1}-x_{k}-x_{*}\right\Vert ^{2}\\
 & =4\left\Vert \Pi_{C}\left(z_{k}-\eta_{k}\nabla f(x_{k},\xi_{k})\right)-\left(\frac{1}{2}x_{k}+\frac{1}{2}x_{*}\right)\right\Vert ^{2}\\
 & =4\left\Vert \Pi_{C}\left(z_{k}-\eta_{k}\nabla f(x_{k},\xi_{k})\right)-\Pi_{C}\left(\frac{1}{2}x_{k}+\frac{1}{2}x_{*}\right)\right\Vert ^{2}\\
 & \leq\left\Vert 2z_{k}-2\eta_{k}\nabla f(x_{k},\xi_{k})-x_{k}-x_{*}\right\Vert ^{2}.
\end{align*}
Now we use $z_{k}=\frac{1}{c_{k}}x_{k}-\left(\frac{1}{c_{k}}-1\right)x_{k-1}$:
\begin{align*}
A_{k+1} & \leq\left\Vert \frac{2}{c_{k}}x_{k}-2\left(\frac{1}{c_{k}}-1\right)x_{k-1}-2\eta_{k}\nabla f(x_{k},\xi_{k})-x_{k}-x_{*}\right\Vert ^{2}\\
 & =\left\Vert 2\left(\frac{1}{c_{k}}-1\right)x_{k}-2\left(\frac{1}{c_{k}}-1\right)x_{k-1}+x_{k}-2\eta_{k}\nabla f(x_{k},\xi_{k})-x_{*}\right\Vert ^{2}\\
 & =\left\Vert x_{k}-2\eta_{k}\nabla f(x_{k},\xi_{k})-x_{*}\right\Vert ^{2}+4\left(\frac{1}{c_{k}}-1\right)^{2}\left\Vert x_{k}-x_{k-1}\right\Vert ^{2}\\
 & +4\left(\frac{1}{c_{k}}-1\right)\left\langle x_{k}-x_{k-1},x_{k}-x_{*}\right\rangle -4\eta_{k}\left(\frac{1}{c_{k}}-1\right)\left\langle \nabla f(x_{k},\xi_{k}),x_{k}-x_{*}\right\rangle .
\end{align*}
Now from Lemma \ref{lem:lyapunov_relation} we have $A_{k}=\left\Vert x_{k}-x_{*}+\lambda_{k}\left(x_{k}-x_{k-1}\right)\right\Vert ^{2}$
thus
\begin{multline}
4\left(\frac{1}{c_{k}}-1\right)\left\langle x_{k}-x_{k-1},x_{k}-x_{*}\right\rangle =\frac{2}{\lambda_{k}}\left(\frac{1}{c_{k}}-1\right)A_{k}  \\
-\frac{2}{\lambda_{k}}\left(\frac{1}{c_{k}}-1\right)\left\Vert x_{k}-x_{*}\right\Vert ^{2}-2\lambda_{k}\left(\frac{1}{c_{k}}-1\right)\left\Vert \left(x_{k}-x_{k-1}\right)\right\Vert ^{2}.
\end{multline}
Notice that:
\begin{align*}
2\frac{1}{\lambda_{k}}\left(\frac{1}{c_{k}}-1\right) & =2\frac{2}{k+1}\left(\frac{k+3}{4}-1\right)\\
 & =\frac{1}{k+1}\left(k+3-4\right)\\
 & =\frac{k-1}{k+1}=\rho_{k}.
\end{align*}
So we have:
\begin{align*}
A_{k+1} & =\left\Vert x_{k}-2\eta_{k}\nabla f(x_{k},\xi_{k})-x_{*}\right\Vert ^{2}+\left(4\left(\frac{1}{c_{k}}-1\right)-2\lambda_{k}\right)\left(\frac{1}{c_{k}}-1\right)\left\Vert x_{k}-x_{k-1}\right\Vert ^{2}\\
 & =\rho_{k}A_{k}-\rho_{k}\left\Vert x_{k}-x_{*}\right\Vert ^{2}-8\eta_{k}\left(\frac{1}{c_{k}}-1\right)\left\langle \nabla f(x_{k},\xi_{k}),x_{k}-x_{k-1}\right\rangle.
\end{align*}
Now note that:
\begin{align*}
4\left(\frac{1}{c_{k}}-1\right)-2\lambda_{k} & =4\left(\frac{k+3}{4}-1\right)-2\frac{k+1}{2}\\
 & =\left(k-1\right)-2\frac{k+1}{2}\\
 & \leq 0.
\end{align*}
Further expanding $\left\Vert x_{k}-2\eta_{k}\nabla f(x_{k},\xi_{k})-x_{*}\right\Vert ^{2}$
and rearranging then gives
\begin{align*}
A_{k+1} & =\rho_{k}A_{k}+\left(1-\rho_{k}\right)\left\Vert x_{k}-x_{*}\right\Vert ^{2}+4\eta_{k}^{2}\left\Vert \nabla f(x_{k},\xi_{k})\right\Vert ^{2}\\
 & =-4\eta_{k}\left\langle \nabla f(x_{k},\xi_{k}),x_{k}-x_{*}\right\rangle -8\eta_{k}\left(\frac{1}{c_{k}}-1\right)\left\langle \nabla f(x_{k},\xi_{k}),x_{k}-x_{k-1}\right\rangle .
\end{align*}
We now apply the two inequalities:
\[
-\left\langle \nabla f\left(x_{k},\xi_{k}\right),x_{k}-x_{*}\right\rangle \leq-\left[f(x_{k},\xi_{k})-f(x_{*},\xi_{k})\right]-\frac{\mu}{2}\left\Vert x_{k}-x_{*}\right\Vert ^{2},
\]
\[
-\left\langle \nabla f\left(x_{k},\xi_{k}\right),x_{k}-x_{k-1}\right\rangle \leq f(x_{k-1},\xi_{k})-f(x_{k},\xi_{k}),
\]
which gives:
\begin{align*}
A_{k+1} & =\rho_{k}A_{k}+\left(1-\rho_{k}-2\mu\eta_{k}\right)\left\Vert x_{k}-x_{*}\right\Vert ^{2}+4\eta_{k}^{2}\left\Vert \nabla f(x_{k},\xi_{k})\right\Vert ^{2}\\
 & =-4\eta_{k}\left[f(x_{k},\xi_{k})-f(x_{*},\xi_{k})\right]+8\eta_{k}\left(\frac{1}{c_{k}}-1\right)\left[f(x_{k-1},\xi_{k})-f(x_{k},\xi_{k})\right].
\end{align*}
Taking expectations and using $\E_{\xi_{k}}\left\Vert \nabla f\left(x_{k},\xi_{k}\right)\right\Vert ^{2}\leq G^{2}$
gives:
\begin{align*}
\E A_{k+1} & =\rho_{k}A_{k}+\left(1-\rho_{k}-2\mu\eta_{k}\right)\left\Vert x_{k}-x_{*}\right\Vert ^{2}+4\eta_{k}^{2}G^{2}\\
 & =-4\eta_{k}\left[f(x_{k})-f(x_{*})\right]+8\eta_{k}\left(\frac{1}{c_{k}}-1\right)\left[f(x_{k-1})-f(x_{k})\right].
\end{align*}
Further grouping of function value terms gives:
\begin{align*}
\E A_{k+1} & =\rho_{k}A_{k}+\left(1-\rho_{k}-2\mu\eta_{k}\right)\left\Vert x_{k}-x_{*}\right\Vert ^{2}+4\eta_{k}^{2}G^{2}\\
 & =-\left(8\eta_{k}\left(\frac{1}{c_{k}}-1\right)+4\eta_{k}\right)\left[f(x_{k})-f(x_{*})\right]+8\eta_{k}\left(\frac{1}{c_{k}}-1\right)\left[f(x_{k-1})-f(x_{*})\right].
\end{align*}
Now we simplify constants, recalling that $\rho_{k}=\frac{k-1}{k+1}$
and $c_{k}=\frac{4}{k+3}$:

\begin{align*}
8\eta_{k}\left(\frac{1}{c_{k}}-1\right) & =2\frac{4}{\mu(k+1)}\left(\frac{k+3}{4}-1\right)\\
 & =\frac{2}{\mu}\frac{1}{k+1}\left(k-1\right)\\
 & =\rho_{k}\frac{2}{\mu},
\end{align*}
using this we have:
\begin{align*}
8\eta_{k}\left(\frac{1}{c_{k}}-1\right)+4\eta_{k} & =\frac{2}{\mu}\frac{k-1}{k+1}+4\frac{1}{\mu(k+1)}\\
 & =\frac{2}{\mu}\frac{k-1+2}{k+1}\\
 & =\frac{2}{\mu}.
\end{align*}
Also note that:
\begin{align*}
1-\rho_{k}-2\mu\eta_{k} & =1-\frac{k-1}{k+1}-\frac{2\mu}{\mu(k+1)}\\
 & =1-\frac{k+1-2}{k+1}-\frac{2}{k+1}\\
 & =0.
\end{align*}
So we have:

\begin{align*}
\E A_{k+1}+\frac{2}{\mu}\left[f(x_{k})-f(x_{*})\right] & =\rho_{k}\left[A_{k}+\frac{2}{\mu}f(x_{k-1})-f(x_{*})\right]+4\eta_{k}^{2}G^{2}.
\end{align*}
Based on the form of this equation, we have a Laypunov function 
\[
B_{k+1}=A_{k+1}+\frac{2}{\mu}\left[f(x_{k})-f(x_{*})\right],
\]
then:
\[
\E B_{k+1}\leq\rho_{k}B_{k}+4\eta_{k}^{2}G^{2},
\]
with $\rho_{k}$ descent plus noise. To finish the proof, we multiply
by $k(k+1)$ and simplify the last term: 
\[
\left(k+1\right)k \E[B_{k+1}]\leq k\left(k-1\right)B_{k}+\frac{4}{\mu^{2}}G^{2}.
\]
We now telescope from $k=1$ to $n$, using the law of total expectation:
\[
\left(n+1\right)n \E[B_{n+1}]\leq\frac{4n}{\mu^{2}}G^{2},
\]
\[
\therefore \E \left[f(x_{n})-f(x_{*})\right]\leq\frac{2G^{2}}{\mu(n+1)}.
\]
\end{proof}

\section{Accelerated Method}
Consider the following iterate averaging form of Nesterov's method
\begin{align}
y_{k} & =\left(1-c_{k+1}\right)x_{k}+c_{k+1}z_{k} \nonumber \\
z_{k+1} & =z_{k}- \rho_k \nabla f(y_{k})\nonumber \\
x_{k+1} & =\left(1-c_{k+1}\right)x_{k}+c_{k+1}z_{k+1}. \label{eq:Nesterov2}
\end{align}
with $z_{0}=x_{0}$. Note the following two key relations, that can
be derived by rearranging the above relations
\begin{equation}
z_{k}=y_{k}-\left(\frac{1}{c_{k+1}}-1\right)\left(x_{k}-y_{k}\right),\label{eq:npropzk}
\end{equation}
and
\begin{equation}
x_{k+1}-y_{k}=c_{k+1}\left(z_{k+1}-z_{k}\right). \label{eq:npropxkp1}
\end{equation}
\begin{lem}
\label{lem:descent-lemma-1} 
Let $f(x,\xi)$ be  $L$--smooth and convex.
 If we set $c_{k+1} = 2/(k+2)$ and $\rho_{k}=(k+1)/(\gamma L)$ then the iterates of  iterate averaging form of Nesterov's method~\eqref{eq:Nesterov2} satisfy
\[
-f(y_{k})\leq-f(x_{k+1})-2L\left(\frac{\gamma}{\left(k+1\right)^{\overline{2}}}-\frac{1}{\left(k+2\right)^{2}}\right)\left\Vert z_{k+1}-z_{k}\right\Vert ^{2}.
\]
\end{lem}
\begin{proof}
We start with the Lipschitz smoothness upper bound:
\[
f(x_{k+1})\leq f(y_{k})+\left\langle \nabla f(y_{k}),x_{k+1}-y_{k}\right\rangle +\frac{L}{2}\left\Vert x_{k+1}-y_{k}\right\Vert ^{2},
\]
\[
\therefore-f(y_{k})\leq-f(x_{k+1})+\left\langle \nabla f(y_{k}),x_{k+1}-y_{k}\right\rangle +\frac{L}{2}\left\Vert x_{k+1}-y_{k}\right\Vert ^{2}.
\]
Using~\eqref{eq:npropxkp1} and
$\nabla f(y_{k})=-L\gamma/(k+1)\left(z_{k+1}-z_{k}\right)$ in the above gives
\[
f(y_{k})\leq-f(x_{k+1})-\frac{L\gamma}{\alpha\left(k+1\right)}\left\langle \left(z_{k+1}-z_{k}\right),c_{k+1}\left(z_{k+1}-z_{k}\right)\right\rangle +\frac{L}{2}\left\Vert c_{k+1}\left(z_{k+1}-z_{k}\right)\right\Vert ^{2}.
\]
Note that $c_{k+1}^{2}=\frac{4}{(k+2)^{2}}$ so:
\[
-f(y_{k})\leq-f(x_{k+1})-\frac{L}{k+1}\frac{2}{(k+2)}\left\Vert z_{k+1}-z_{k}\right\Vert ^{2}+\frac{L}{2}\frac{4}{(k+2)^{2}}\left\Vert z_{k+1}-z_{k}\right\Vert ^{2}.
\]
Grouping terms gives the lemma.
\end{proof}

\subsubsection*{Proof of Theorem \ref{thm:nesterov-kplus1sqr}}
\begin{thm}
Let $f(x,\xi)$ be  $L$--smooth and convex.
Let $x_k$ be given by the iterate averaging form of Nesterov's method~\eqref{eq:Nesterov2}.
  If we set $c_{k+1} = 2/(k+2)$ and $\rho_{k}=(k+1)/(\gamma L)$ with $\gamma=2$ then
\begin{equation}
f(x_{n})-f(x_{*})	\leq\frac{2L}{n^{\overline{2}}}\left\Vert x_{0}-x_{*}\right\Vert ^{2}.
\end{equation}
\end{thm}
\begin{proof}
We start by expanding a distance to solution term:
\begin{align*}
\left\Vert z_{k+1}-x_{*}\right\Vert ^{2} & =\left\Vert z_{k}-x_{*}-\left(z_{k}-z_{k+1}\right)\right\Vert ^{2}\\
 & =\left\Vert z_{k}-x_{*}\right\Vert ^{2}-2(k+1)\frac{1}{\gamma L}\left\langle \nabla f(y_{k}),z_{k}-x_{*}\right\rangle +\left\Vert z_{k+1}-z_{k}\right\Vert ^{2}.
\end{align*}
Simplifying the inner product term:
\begin{align*}
-2(k+1)\frac{1}{\gamma L}\left\langle \nabla f(y_{k}),z_{k}-x_{*}\right\rangle  & \overset{\eqref{eq:npropzk}}{=}-2(k+1)\frac{1}{\gamma L}\left\langle \nabla f(y_{k}),y_{k}-\left(\frac{1}{c_{k+1}}-1\right)\left(x_{k}-y_{k}\right)-x_{*}\right\rangle \\
 & \,=-2(k+1)\frac{1}{\gamma L}\left\langle \nabla f(y_{k}),y_{k}-x_{*}\right\rangle \\
 & -2(k+1)\frac{1}{\gamma L}\left(\frac{1}{c_{k+1}}-1\right)\left\langle \nabla f(y_{k}),y_{k}-x_{k}.\right\rangle 
\end{align*}
Then we apply the inequalities:
\[
-\left\langle \nabla f\left(y_{k}\right),y_{k}-x_{*}\right\rangle \leq f(x_{*})-f(y_{k}),
\]
\[
-\left\langle \nabla f\left(y_{k}\right),y_{k}-x_{k}\right\rangle \leq f(x_{k})-f(y_{k}).
\]
So we have
\begin{align*}
\left\Vert z_{k+1}-x_{*}\right\Vert ^{2} & \leq\left\Vert z_{k}-x_{*}\right\Vert ^{2}+\left\Vert z_{k+1}-z_{k}\right\Vert ^{2}\\
 & -2(k+1)\frac{1}{\gamma L}\left[f(y_{k})-f(x_{*})\right]+2(k+1)\frac{1}{\gamma L}\left(\frac{1}{c_{k+1}}-1\right)\left[f(x_{k})-f(y_{k})\right].
\end{align*}
Now rearranging the function value terms and using that $c_{k+1} = \frac{2}{k+2}$ gives
\begin{align*}
\left\Vert z_{k+1}-x_{*}\right\Vert ^{2} & \leq\left\Vert z_{k}-x_{*}\right\Vert ^{2}+\left\Vert z_{k+1}-z_{k}\right\Vert ^{2}\\
 & -(k+1)^{\overline{2}}\frac{1}{\gamma L}\left[f(y_{k})-f(x_{*})\right]+(k+1)^{\overline{2}}\frac{1}{\gamma L}\left(\frac{1}{c_{k+1}}-1\right)\left[f(x_{k})-f(x_{*})\right].
\end{align*}
Now we use Lemma \ref{lem:descent-lemma-1} on $-f(y_{k})$ gives
\[
-(k+1)^{\overline{2}}\frac{1}{\gamma L}f(y_{k})\leq-(k+1)^{\overline{2}}\frac{1}{\gamma L}f(x_{k+1})-2\left(1-\frac{(k+1)}{\gamma(k+2)}\right)\left\Vert z_{k+1}-z_{k}\right\Vert ^{2},
\]
which combined with the preceding result gives
\begin{align*}
\left\Vert z_{k+1}-x_{*}\right\Vert ^{2} & \leq\left\Vert z_{k}-x_{*}\right\Vert ^{2}+\left(1-2\left(1-\frac{k+1}{\gamma(k+2)}\right)\right)\left\Vert z_{k+1}-z_{k}\right\Vert ^{2}\\
 & -(k+1)^{\overline{2}}\frac{1}{\gamma L}\left[f(x_{k+1})-f(x_{*})\right]+(k+1)^{\overline{2}}\frac{1}{\gamma L}\left(\frac{1}{c_{k+1}}-1\right)\left[f(x_{k})-f(x_{*})\right].
\end{align*}
When $\gamma=2$ then $-2\left(1-\frac{k+1}{\gamma(k+2)}\right)\leq-1$ so
\begin{align*}
\left\Vert z_{k+1}-x_{*}\right\Vert ^{2}+&(k+1)^{\overline{2}}\frac{1}{\gamma L}\left[f(x_{k+1})-f(x_{*})\right]  \\
&\leq\left\Vert z_{k}-x_{*}\right\Vert ^{2}+(k+1)^{\overline{2}}\frac{1}{\gamma L}\left(\frac{1}{c_{k+1}}-1\right)\left[f(x_{k})-f(x_{*})\right].
\end{align*}
Now we apply Lemma \ref{lem:ck_iterateweighting} to give a telescopable
sum:
\begin{align*}
\left\Vert z_{k+1}-x_{*}\right\Vert ^{2}+(k+1)^{\overline{2}}\frac{1}{\gamma L}\left[f(x_{k+1})-f(x_{*})\right] & \leq\left\Vert z_{k}-x_{*}\right\Vert ^{2}+k^{\overline{2}}\frac{1}{\gamma L}\left[f(x_{k})-f(x_{*})\right].
\end{align*}
After telescoping:
\begin{align*}
f(x_{n})-f(x_{*}) & \leq\frac{2L}{n^{\overline{2}}}\left\Vert x_{0}-x_{*}\right\Vert ^{2}.
\end{align*}
\end{proof}

\section{SVRGM}
\begin{lem}
\label{lem:svrg-noise-bound} \citep{svrg} The following bound holds
for $g_{t}^{s}$ at each step:
\[
\E \left\Vert g_{t}^{s}\right\Vert ^{2}\leq4L\left[f\left(x_{t}^{s}\right)-f\left(x_{*}\right)\right]+4L\left[f\left(\tilde{x}^{s-1}\right)-f\left(x_{*}\right)\right].
\]
\end{lem}

\subsection{Proof of Theorem~\ref{thm:SVRGMconvex} (Convex Case) }
\begin{thm}
At the end of epoch $S$, 
when using $r=1/2$ factorial power momentum given by
\[c_{t}=\frac{1/2+1}{t+1/2+1},\]
and
step size $\eta=\frac{1}{6L}$, the expected function value is bounded
by:
\[
\E \left[f(x_{m_{s}-1}^{S})-f(x_{*})\right]\leq\frac{1}{2^{S}}\left[f\left(x_{0}\right)-f\left(x_{*}\right)\right]+\frac{9L\left\Vert x_{0}-x_{*}\right\Vert ^{2}}{2^{S}m_{0}}.
\]
\end{thm}

\begin{proof}
We start in the same fashion as for non-variance reduced momentum
methods:
\begin{align*}
\E \left\Vert z_{t+1}^{s}-x_{*}\right\Vert ^{2} & =\E \left\Vert z_{t}^{s}-\eta g_{t}^{s}-x_{*}\right\Vert ^{2}\\
 & =\left\Vert z_{t}^{s}-x_{*}\right\Vert ^{2}-2\eta\left\langle \nabla f(x_{t}^{s}),z_{t}^{s}-x_{*}\right\rangle +\eta^{2}\E \left\Vert g_{t}^{s}\right\Vert ^{2}\\
 & =\left\Vert z_{t}^{s}-x_{*}\right\Vert ^{2}-2\eta_{t}\left\langle \nabla f(x_{t}^{s}),x_{t}^{s}-\left(\frac{1}{c_{t}}-1\right)\left(x_{t-1}^{s}-x_{t}^{s}\right)-x_{*}\right\rangle +\eta^{2}\E \left\Vert g_{t}^{s}\right\Vert ^{2}\\
 & =\left\Vert z_{t}^{s}-x_{*}\right\Vert ^{2}+\eta^{2}\E \left\Vert g_{t}^{s}\right\Vert ^{2}\\
 & -2\eta\left\langle \nabla f(x_{t}^{s}),x_{t}^{s}-x_{*}\right\rangle -2\eta\left(\frac{1}{c_{t}}-1\right)\left\langle \nabla f(x_{t}^{s}),x_{t}^{s}-x_{t-1}^{s}\right\rangle .
\end{align*}
Using the following two convexity inequalities
\[
\left\langle \nabla f\left(x_{t}^{s}\right),x_{*}-x_{t}^{s}\right\rangle \leq f(x_{*})-f(x_{t}^{s}),
\]
\[
\left\langle \nabla f\left(x_{t}^{s}\right),x_{t-1}^{s}-x_{t}^{s}\right\rangle \leq f(x_{t-1}^{s})-f(x_{t}^{s}),
\]
combined with $(1/c_{t}-1)\geq0$ gives 
\begin{align*}
\E \left\Vert z_{t+1}^{s}-x_{*}\right\Vert ^{2} & \leq\left\Vert z_{t}^{s}-z_{*}\right\Vert ^{2}+\eta^{2}\E \left\Vert g_{t}^{s}\right\Vert ^{2}\\
 & -2\eta\left[f(x_{t}^{s})-f(x_{*})\right]-2\eta\left(\frac{1}{c_{t}}-1\right)\left[f(x_{t}^{s})-f(x_{t-1}^{s})\right].
\end{align*}
Now rearranging further:
\begin{align*}
\E \left\Vert z_{t+1}^{s}-x_{*}\right\Vert ^{2} & \leq\left\Vert z_{t}^{s}-z_{*}\right\Vert ^{2}+\eta^{2}\E \left\Vert g_{t}^{s}\right\Vert ^{2}\\
 & -2\eta\frac{1}{c_{t}}\left[f(x_{t}^{s})-f(x_{*})\right]+2\eta\left(\frac{1}{c_{t}}-1\right)\left[f(x_{t-1}^{s})-f(x_{*})\right].
\end{align*}
Now using Lemma\ref{lem:svrg-noise-bound}
\begin{align*}
\E \left\Vert z_{t+1}^{s}-x_{*}\right\Vert ^{2} & \leq\left\Vert z_{t}^{s}-z_{*}\right\Vert ^{2}+4L\eta^{2}\left[f\left(\tilde{x}^{s-1}\right)-f\left(x_{*}\right)\right]\\
 & -2\eta\left(\frac{1}{c_{t}}-2\eta L\right)\left[f(x_{t}^{s})-f(x_{*})\right]+2\eta\left(\frac{1}{c_{t}}-1\right)\left[f(x_{t-1}^{s})-f(x_{*})\right].
\end{align*}
Now for the purposes of telescoping, define $\lambda_{t}=p(t+1)$,
we want to choose $p>0$ such that
\[
\frac{1}{c_{t}}-2\eta L=p(t+1) \quad \mbox{and} \quad  \frac{1}{c_{t}}-1=pt.
\]
These equations are satisfied for $p=1-2L\eta=\frac{2}{3}$, when
$\eta=\frac{1}{6L}$ and 
\[
c_{t}=\frac{1}{pt+1}=\frac{1/2+1}{t+1/2+1}.
\]
This corresponds to $r=1/2$ factorial power momentum. So we have:%

\begin{align*}
\E \left\Vert z_{t+1}^{s}-x_{*}\right\Vert ^{2} & \leq\left\Vert z_{t}^{s}-z_{*}\right\Vert ^{2}+\frac{1}{9L}\left[f\left(\tilde{x}^{s-1}\right)-f\left(x_{*}\right)\right]\\
 & -\frac{2}{9L}(t+1)\left[f(x_{t}^{s})-f(x_{*})\right]+\frac{2}{9L}t\left[f(x_{t-1}^{s})-f(x_{*})\right].
\end{align*}
We now telescope from $t=0$ to $t=m_{s}-1$, using the law of total
expectation (i.e. $\E \left[\E \left[X|Y\right]\right]=\E [X]$), so that
this expectation is unconditional:
\begin{align*}
\E\left\Vert z_{m_{s}-1}^{s}-x_{*}\right\Vert ^{2} & \leq\left\Vert z_{0}^{s}-z_{*}\right\Vert ^{2}+\frac{m_{s}}{9L}\left[f\left(\tilde{x}^{s-1}\right)-f\left(x_{*}\right)\right] -\frac{2m_{s}}{9L}\left[f(x_{t}^{s})-f(x_{*})\right].
\end{align*}
Which we can write as:
\begin{align*}
\frac{9L}{2m_{s}}\E\left\Vert z_{m_{s}-1}^{s}-x_{*}\right\Vert ^{2}+\left[f(x_{t}^{s})-f(x_{*})\right] & \leq\frac{9L}{2m_{s}}\left\Vert z_{0}^{s}-z_{*}\right\Vert ^{2}+\frac{1}{2}\left[f\left(\tilde{x}^{s-1}\right)-f\left(x_{*}\right)\right]
\end{align*}
Noting that the choice $z_{0}^{s}=z_{m_{s-1}-1}^{s-1}$ and $m_{s}=2m_{s-1}$
gives:
\[
\frac{\left\Vert z_{0}^{s}-x_{*}\right\Vert ^{2}}{m_{s}}=\frac{1}{2}\frac{\left\Vert z_{m_{s-1}-1}^{s-1}-x_{*}\right\Vert ^{2}}{m_{s-1}}
\]
So we may form the Lyapunov function:
\[
B^{s}=\frac{9L}{2m_{s}}\E\left\Vert z_{m_{s}-1}^{s}-x_{*}\right\Vert ^{2}+\left[f(x_{t}^{s})-f(x_{*})\right]
\]
which gives the simple relation:
\[
\E\left[B^{s}\right]\leq\frac{1}{2}\E\left[B^{s-1}\right].
\]
So after $S$ epochs we have:
\[
\E\left[B^{s}\right]\leq2^{-S}B^{0}.
\]
and so:
\[
\E\left[f(x_{m_{s}-1}^{S})-f(x_{*})\right]\leq\frac{1}{2^{S}}\left[f\left(x_{0}\right)-f\left(x_{*}\right)\right]+\frac{9L\left\Vert x_{0}-x_{*}\right\Vert ^{2}}{2^{S}m_{0}}
\]
\end{proof}

\subsection{Proof of Theorem~\ref{thm:SVRGMstrconvex} (Strongly Convex Case)}
\begin{thm}
When each $f_{i}$ is strongly convex with constant $\mu$, we may
use $m,c,\eta$ constants that don't depend on the step. In particular,
after epoch $s$, when $m=6\kappa$ and $c=\frac{5}{3}\frac{1}{4\kappa+1},$
and $\eta=1/(10L)$:
\[
\E\left[B^{s}\right]\leq\frac{6}{10}B^{s-1},
\]
where:
\[
B^{s}=\E\left[f(\tilde{x}^{s})-f(x_{*})\right]+\frac{3}{4}\mu\left\Vert x_{m_{s}}^{s}-x_{*}+\lambda\left(x_{m_{s}}^{s}-x_{m_{s}-1}^{s}\right)\right\Vert ^{2}.
\]
\end{thm}

\begin{proof}
We can use the same proof technique as we applied in the non-variance
reduced case to deduce the following 1-step bound:
\begin{align*}
\E A_{t+1}^{s} & \leq\left(1-\rho-\mu\nu\right)\left\Vert x_{t}^{s}-x_{*}\right\Vert ^{2}
  +\rho A_{t}+4L\nu^{2}\left[f(\tilde{x}^{s-1})-f(x_{*})\right]\\
 & -2\nu\left(1+\rho\lambda-2L\nu\right)\left[f(x_{t}^{s})-f(x_{*})\right]+2\rho\lambda\nu\left[f(x_{t-1}^{s})-f(x_{*})\right]
\end{align*}
Where
\[
\rho=\frac{\left(\lambda+1\right)\beta}{\lambda} \quad \mbox{and} \quad
\nu=\left(\lambda+1\right)\alpha.
\]
We need $1-\rho-\mu\nu\leq0$, which suggests for step sizes of the
form $\nu=1/\left(qL\right),$
\[
\rho=1-\mu\nu=1-\frac{1}{q\kappa}.
\]
Now in order to see a $\rho$ decrease in function value each step,
we will require:
\[
-2\nu\left(1+\rho\lambda-2L\nu\right)\leq-2\lambda\nu,
\]
so solving at equality gives
\[
1+\rho\lambda-2L\nu=\lambda,
\]
\[
\therefore1-2L\nu=\left(1-\rho\right)\lambda,
\]
\[
\lambda=\frac{1-2/q}{1/q\kappa}=\left(q-2\right)\kappa.
\]
This gives:
\begin{align*}
2\lambda\nu & =2\left(q-2\right)\kappa\frac{1}{qL}
  \; =\frac{2}{\mu}\left(1-\frac{2}{q}\right)
\end{align*}
Making these substitutions, our one-step bound can be written as:
\begin{align*}
\E A_{t+1}^{s}+\frac{2}{\mu}\left(1-\frac{2}{q}\right)\left[f(x_{t}^{s})-f(x_{*})\right] & \leq\rho A_{t}^{s}+\rho\frac{2}{\mu}\left(1-\frac{2}{q}\right)\left[f(x_{t-1}^{s})-f(x_{*})\right]\\
 & +\frac{4}{q^{2}L}\left[f(\tilde{x}^{s-1})-f(x_{*})\right].
\end{align*}
We can now telescope using the sum of a geometric series $\sum_{i=0}^{k-1}\rho^{i}=\frac{1-\rho^{k}}{1-\rho}$
and the law of total expectation to give:
\begin{align*}
\E A_{m+1}^{s}+\frac{2}{\mu}\left(1-\frac{2}{q}\right)\left[f(x_{m}^{s})-f(x_{*})\right] & \leq\rho^{m}A_{0}^{s}+\rho^{m}\frac{2}{\mu}\left(1-\frac{2}{q}\right)\left[f(\tilde{x}^{s-1})-f(x_{*})\right]\\
 & +\frac{1-\rho^{m}}{1-\rho}\frac{4}{q^{2}L}\left[f(\tilde{x}^{s-1})-f(x_{*})\right].
\end{align*}
These expectations are now unconditional. Now multiplying by $\mu/2$,
simplifying with $1-\rho=\frac{1}{q\kappa}$ gives:
\begin{multline*}
\frac{\mu}{2}\E A_{m+1}^{s}+\left(1-\frac{2}{q}\right)\left[f(x_{m}^{s})-f(x_{*})\right] 
 \leq \rho^{m}\frac{\mu}{2}A_{0}^{s}+\left(\rho^{m}\left(1-\frac{2}{q}\right)+\frac{2}{q}\left(1-\rho^{m}\right)\right)\left[f(\tilde{x}^{s-1})-f(x_{*})\right].
\end{multline*}
Dividing by $\left(1-\frac{2}{q}\right)$:
\begin{multline*}
\frac{\mu}{2}\frac{q}{q-2}EA_{m+1}^{s}+\left[f(x_{m}^{s})-f(x_{*})\right] 
 \leq\rho^{m}\frac{\mu}{2}\frac{q}{q-2}A_{0}^{s}+\left(\rho^{m}+\frac{2}{q-2}\left(1-\rho^{m}\right)\right)\left[f(\tilde{x}^{s-1})-f(x_{*})\right]
\end{multline*}
Now we can try $q=6$ for instance, giving
\[
\rho^{m}+\frac{2}{q-2}\left(1-\rho^{m}\right)=\rho^{m}+\frac{1}{2}\left(1-\rho^{m}\right)=\frac{1}{2}\rho^{m}+\frac{1}{2}
\]
Then if we use $m=6\kappa$ we get $\rho^{m}\leq\exp(-1)\leq2/5$
for $m=6$ to give:
\begin{align*}
\frac{3}{4}\mu \E A_{m+1}^{s}+\left[f(x_{m}^{s})-f(x_{*})\right] & \leq\frac{6}{10}\left[\frac{3}{4}\mu A_{0}^{s}+\left[f(\tilde{x}^{s-1})-f(x_{*})\right]\right]
\end{align*}
Then we may determine the momentum and step size constants $\alpha,\beta:$
\begin{align*}
\beta & =\frac{\lambda}{\lambda+1}\rho=\frac{\left(6-2\right)\kappa}{\left(6-2\right)\kappa+1}\left(1-\frac{1}{6\kappa}\right)\\
 & =\frac{4\kappa}{4\kappa+1}\left(\frac{6\kappa-1}{6\kappa}\right)\\
 & =\frac{2}{3}\frac{6\kappa-1}{4\kappa+1}\\
 & =\frac{4\kappa-2/3}{4\kappa+1}\\
 & =1-\frac{5/3}{4\kappa+1}
\end{align*}
and
\[
\alpha=\frac{\nu}{\lambda+1}=\frac{1}{6L}\frac{1}{4\kappa+1}.
\]
To write in iterate averaging form, we have $\beta=1-c$ and 
\[
c=\frac{5}{3}\frac{1}{4\kappa+1},
\]
from $\alpha_{k}=\eta c$ we get for $\eta$ that
\[
\eta=\frac{\frac{1}{6L}\frac{1}{4\kappa+1}}{\frac{5}{3}\frac{1}{4\kappa+1}}=\frac{1}{10L}.
\]
\end{proof}

\section{Dual averaging}
First we provide a convergence theorem for the dual averaging method that does not use factorial powers to set the $\hat{\beta}_{k}$ parameters.
\begin{thm} \label{thm:dualaver}
Let
\[
\delta_n = \max_{x,\left\Vert x\right\Vert \leq R}\left\{\sum_{i=0}^{n}\left\langle \nabla f\left(x_{i}\right),x_{i}-x\right\rangle \right\}.
\]
Consider the Dual Averaging method
\begin{align}
s_{k+1} &= s_{k}+\nabla f\left(x_{k}\right)\nonumber \\
x_{k+1}  &= \arg\min_{x}\left\{ \left\langle s_{k+1},x\right\rangle +\hat{\beta}_{k+1}\frac{\gamma}{2}\left\Vert x-x_{0}\right\Vert ^{2}\right\} ,\label{eq:dual-average2}
\end{align}
where the sequence $\hat{\beta}_{k}$ is defined recursively by 
\begin{equation}\label{eq:betas}
 \hat{\beta}_{0}=\hat{\beta}_{1}=1\mbox{, and }
\hat{\beta}_{k+1}=\hat{\beta}_{k+1}+1/\hat{\beta}_{k+1}.
\end{equation}
If $\gamma=\frac{G}{\sqrt{2}R}$ then
\[
\frac{1}{k+1}\delta_{k+1}\leq\left(\frac{\sqrt{2}}{\left(1+\sqrt{3}\right)}\frac{1}{(k+1)}+\frac{2}{\sqrt{k+1}}\right)RG.
\]
\end{thm}
\begin{proof}
\citet{nes-da} establishes the following bound:
\[
\delta_{k}\leq\gamma\hat{\beta}_{k+1}R^{2}+\frac{1}{2}G^{2}\frac{1}{\gamma}\sum_{i=0}^{k}\frac{1}{\hat{\beta}_{i}}.
\]
The $\hat{\beta}_{i}$ sequence given in \citet{nes-da} satisfies $\sum_{i=0}^{k}\frac{1}{\hat{\beta}_{i}}=\hat{\beta}_{k+1}$ and $\hat{\beta}_{k+1} \leq \frac{1}{1+\sqrt{3}}+\sqrt{2k+1}$ so we have:
\[
\delta_{k}\leq\left(\gamma R^{2}+\frac{1}{2}G^{2}\frac{1}{\gamma}\right)\left(\frac{1}{1+\sqrt{3}}+\sqrt{2k+1}\right).
\]
The optimal step size is $\gamma=\frac{G}{\sqrt{2}R}$ So:
\[
\delta_{k}\leq\left(\frac{1}{\sqrt{2}}RG+\frac{1}{\sqrt{2}}RG\right)\left(\frac{1}{1+\sqrt{3}}+\sqrt{2k+1}\right)
\]
\[
\delta_{k}\leq RG\left(\frac{\sqrt{2}}{1+\sqrt{3}}+\sqrt{4k+2}\right).
\]
Using the concavity of the square-root function:
\begin{align*}
\sqrt{4k+2} & \leq\sqrt{4k+4}+\frac{1}{2}\frac{4k+2-4k-4}{\sqrt{4k+4}}\\
 & =\sqrt{4k+4}-\frac{-1}{\sqrt{4k+4}}.
\end{align*}
We need to normalize this quantity by $1/(k+1)$, so we have:
\begin{align*}
\frac{\sqrt{4k+2}}{k+1} & \leq\frac{\sqrt{4k+4}}{k+1}-\frac{1}{2\left(k+1\right)^{3/2}}\\
 & \leq \frac{2}{\sqrt{k+1}}.
\end{align*}
Therefore the bound on the normalization of $\delta$ is:
\[
\frac{1}{k+1}\delta_{k+1}\leq\left(\frac{\sqrt{2}}{\left(1+\sqrt{3}\right)}\frac{1}{(k+1)}+\frac{2}{\sqrt{k+1}}\right)RG.
\]
\end{proof}

Next we show how using factorial powers to set the $\hat{\beta}_k$ parameters can result in a tighter analysis and a simple proof.

\subsection*{Proof of Theorem~\ref{thm:dualaver-main}}
\begin{thm}
after $n$ steps of the dual averaging method with $\hat{\beta}_{k}=1/\left(k+1\right)^{\overline{-1/2}}$ and $\gamma=G/R$:
\[
\frac{1}{k+1}\delta_{k+1}  \leq 2RG(n+2)^{\overline{-1/2}}<\frac{2RG}{\sqrt{n+1}}.
\]
\end{thm}
\begin{proof}
Recall the bound:
\[
\delta_{k}\leq\gamma\hat{\beta}_{k+1}R^{2}+\frac{1}{2}G^{2}\frac{1}{\gamma}\sum_{i=0}^{k}\frac{1}{\hat{\beta}_{i}}.
\]
We use $\hat{\beta}_{i}=1/\left(i+1\right)^{\overline{-1/2}}$ the sum is:
\begin{align*}
\sum_{i=0}^{k}\frac{1}{\hat{\beta}_{i}} & =\frac{1}{1-1/2}\left(k+1\right)^{\overline{1/2}}-\frac{1}{1-1/2}\left(1\right)^{\overline{1/2}}.
\end{align*}
Recall also that:
\[
\hat{\beta}_{k+1}=\frac{1}{\left(k+2\right)^{\overline{-1/2}}}=\left(k+3/2\right)^{\overline{1/2}}.
\]
So:
\[
\delta_{k}\leq\gamma R^{2}\left(k+3/2\right)^{\overline{1/2}}+G^{2}\left(\left(k+1\right)^{\overline{1/2}}-2\left(1\right)^{\overline{1/2}}\right).
\]
Using step size $\gamma=G/R$:
\begin{align*}
\delta_{k} & \leq RG\left(k+3/2\right)^{\overline{1/2}}+RG\left(\left(k+1\right)^{\overline{1/2}}-2\left(1\right)^{\overline{1/2}}\right)\\
 & =RG\left(\left(k+3/2\right)^{\overline{1/2}}+\left(k+1\right)^{\overline{1/2}}-2\left(1\right)^{\overline{1/2}}\right)\\
 & \leq2RG\left(k+1\right)^{\overline{1/2}}.
\end{align*}
Now to normalize by $1/(k+1)$ we use:
\[
\frac{\left(k+1\right)^{\overline{r+q}}}{\left(k+1\right)^{\overline{r}}}=\left(k+1+r\right)^{\overline{q}},
\]
with $r=1$ and $q=-1/2$, so that:
\[
\frac{\left(k+1\right)^{\overline{1/2}}}{k+1}=\left(k+2\right)^{\overline{-1/2}}.
\]
We further use $(k+2)^{\overline{-1/2}}<(k+1)^{-1/2}$, giving:
\[
\frac{1}{k+1}\delta_{k}<\frac{2RG}{\sqrt{k+1}}.
\]
\end{proof}

\section{SVRGM Experiments}
\begin{figure}
\includegraphics[width=\textwidth]{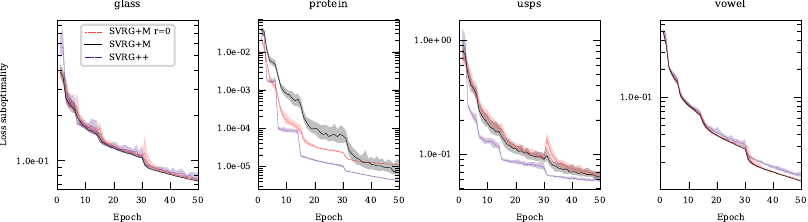}
\caption{\label{plt:svrg+m}SVRGM training loss convergence}
\end{figure}
We compared the SVRGM method against SVRG both with the $r=1/2$ momentum suggested by the theory as well as equal weighted momentum. We used the same test setup as for our \SGDM experiments, except without the addition of weight decay in order to test the non-strongly convex convergence. Since the selection of step-size is less clear in the non-strongly convex case, here we used a step-size sweep on a power-of-2 grid, and we reported the results of the best step-size for each method. As shown in Figure~\ref{plt:svrg+m}, SVRGM is faster on two of the test problems and slower on two. The flat momentum variant is a little slower than $r=1/2$ momentum, however not significantly so.

\end{document}